\documentclass[preprint,12pt,english]{elsarticle}

\usepackage{multicol}
\usepackage{amsfonts}
\usepackage{subfigure}
\usepackage{multirow}
\usepackage{setspace}
\usepackage{graphicx}
\usepackage{color}
\usepackage{threeparttable}
\usepackage{pst-all}
\usepackage{tabularx}
\usepackage{babel}
\usepackage{algorithm}
\usepackage{algpseudocode}
\usepackage{graphics}
\usepackage{epsfig}
\usepackage{amssymb}  % assumes amsmath package installed
\usepackage{amsmath} % assumes amsmath package installed
\usepackage{geometry}
\usepackage[colorlinks,linkcolor=red,urlcolor=blue]{hyperref}
\usepackage{bm}
\usepackage{url}
\usepackage{array}
\usepackage{makecell}
\journal{Computer Vision and Image Understanding}

\begin{document}
\begin{spacing}{1.25}

\begin{frontmatter}



\title{Face Alignment In-the-Wild: A Survey}


\author[adress1,adress2]{Xin~Jin}

\author[adress1,adress2]{Xiaoyang~Tan \corref{cort}}
\cortext[cort]{Corresponding author:
  Tel.: +86-25-8489-6490/6491 (Ext 12106 ) (O);
  fax:  +86-25-8489-2452;
  E-mail:x.tan@nuaa.edu.cn.}

\address[adress1]{Department of Computer Science and Engineering, Nanjing University of Aeronautics and Astronautics, \#29 Yudao Street, Nanjing 210016, P.R. China}
\address[adress2]{Collaborative Innovation Center of Novel Software Technology and Industrialization, Nanjing University of Aeronautics and Astronautics, Nanjing 210016, China}

\begin{abstract}
Over the last two decades, face alignment or localizing fiducial facial points has received increasing attention owing to its comprehensive applications in automatic face analysis. However, such a task has proven extremely challenging in unconstrained environments due to many confounding factors, such as pose, occlusions, expression and illumination. While numerous techniques have been developed to address these challenges, this problem is still far away from being solved. In this survey, we present an up-to-date critical review of the existing literatures on face alignment, focusing on those methods addressing overall difficulties and challenges of this topic under uncontrolled conditions. Specifically, we categorize existing face alignment techniques, present detailed descriptions of the prominent algorithms within each category, and discuss their advantages and disadvantages. Furthermore, we organize special discussions on the practical aspects of face alignment \emph{in-the-wild}, towards the development of a robust face alignment system. In addition, we show performance statistics of the state of the art, and conclude this paper with several promising directions for future research.
\end{abstract}

\begin{keyword}
Face alignment, Active appearance model, Constrained local model, Cascaded regression, Deep convolutional neural networks.
\end{keyword}

\end{frontmatter}

\section{Introduction}\label{sec_introduction}
Fiducial facial points refer to the predefined landmarks on a face graph, which are mainly located around or centered at the facial components such as eyes, mouth, nose and chin (see Fig. \ref{fig_intro_landmark}). Localizing these facial points, which is also known as face alignment, has recently received significant attention in computer vision, especially during the last decade. At least two reasons account for this. Firstly, many important tasks, such as face recognition, face tracking, facial expression recognition, head pose estimation, can benefit from precise facial point localization. Secondly, although some level of success has been achieved in recent years, face alignment in unconstrained environments is so challenging that it remains an open problem in computer vision, and continues to attract researchers to attack it.

While face detection is generally regarded as the starting point for all face analysis tasks \cite{zafeiriou2015survey}, face alignment can be regarded as \emph{an important and essential intermediary step} for many subsequent face analyses that range from biometric recognition to mental state understanding. Concrete tasks may differ in the number and type of the needed facial points, as well as the way these points are used. Below we give some details on three typical tasks where face alignment plays a prominent role:
\begin{itemize}
\item \emph{Face recognition:} Face alignment is widely used by face recognition algorithms to improve their robustness against pose variations. For example, in the stage of face registration, the first step is usually to locate some major facial points and use them as anchor points for affine warping, while other face recognition algorithms, such as feature-based (structural) matching \cite{zhao2003face,campadelli2003feature}, rely on accurate face alignment to build the correspondence among local features (e.g, eyes, nose, mouth, etc.) to be matched.
\begin{figure}[tb]
\centering
\includegraphics[width=0.48\textwidth]{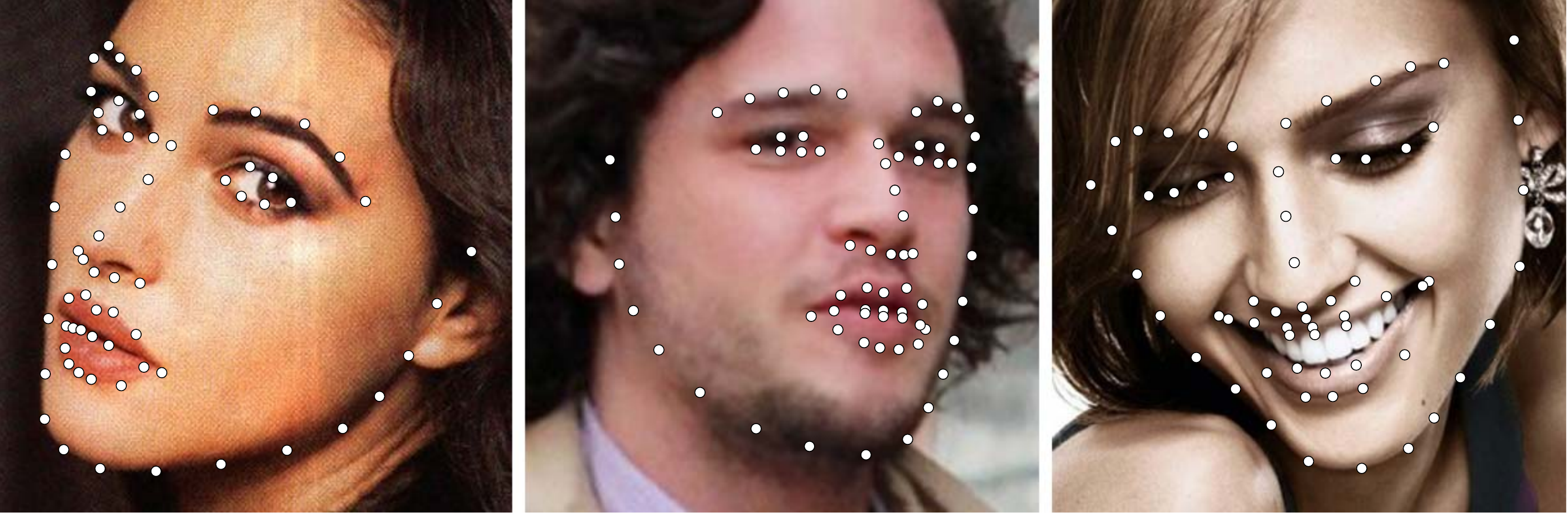}\label{fig_intro_landmark}
\caption{Illustration of some example face images with 68 manually annotated points from the IBUG database \cite{sagonas2013300}.}
\end{figure}
\item \emph{Attribute computing:}  Face alignment is also beneficial to  facial attribute computing, since many facial attributes such as eyeglasses and nose shape are closely related to specific spatial positions of a face. In \cite{kumar2009attribute}, six facial points are localized to compute qualitative attributes and similes that are then used for robust face verification in unconstrained conditions.
\begin{figure*}[!htb]
\centerline{
\subfigure[Pose]{\includegraphics[width=0.24\textwidth]{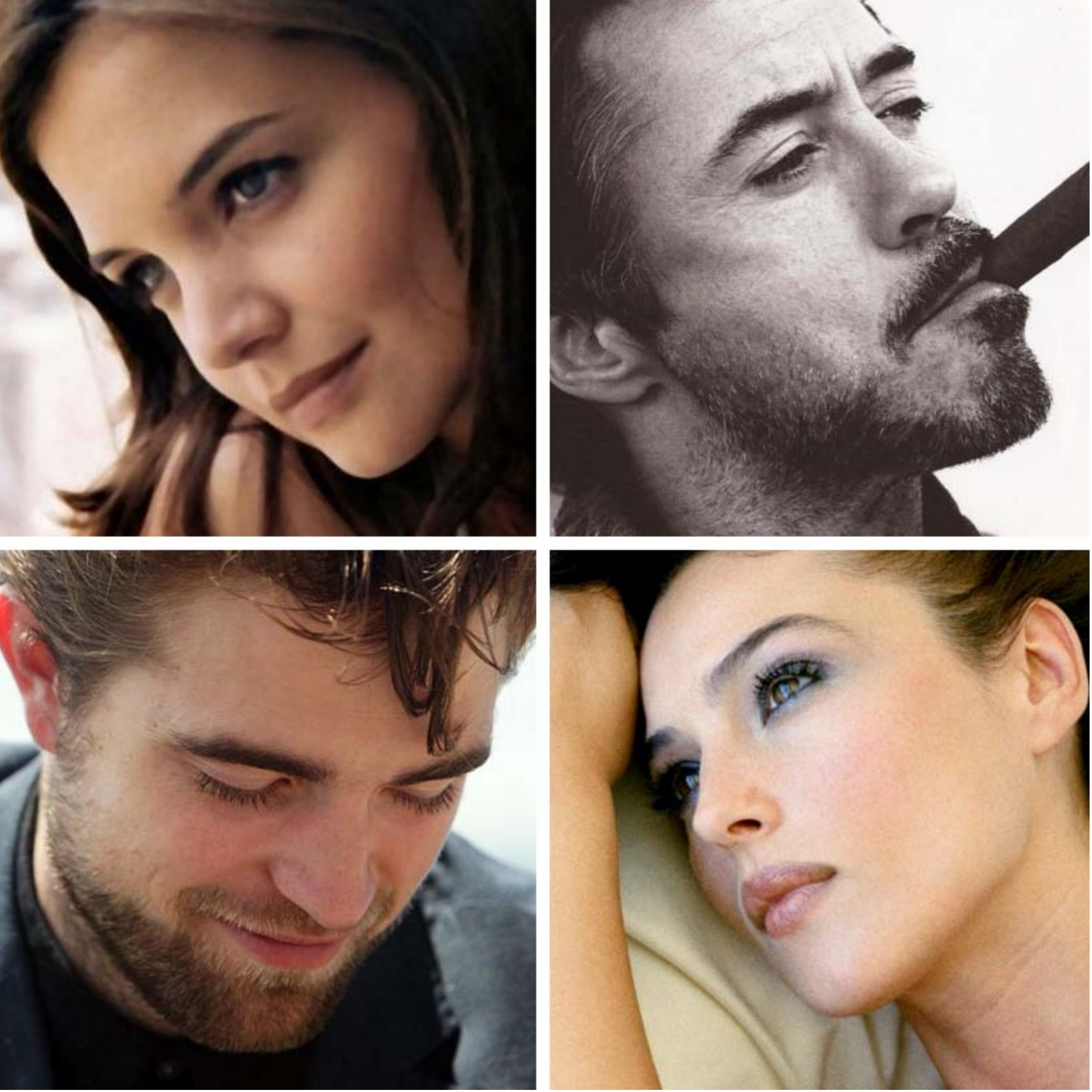}\label{fig_intro_challenge_pose}}
\subfigure[Occlusion]{\includegraphics[width=0.24\textwidth]{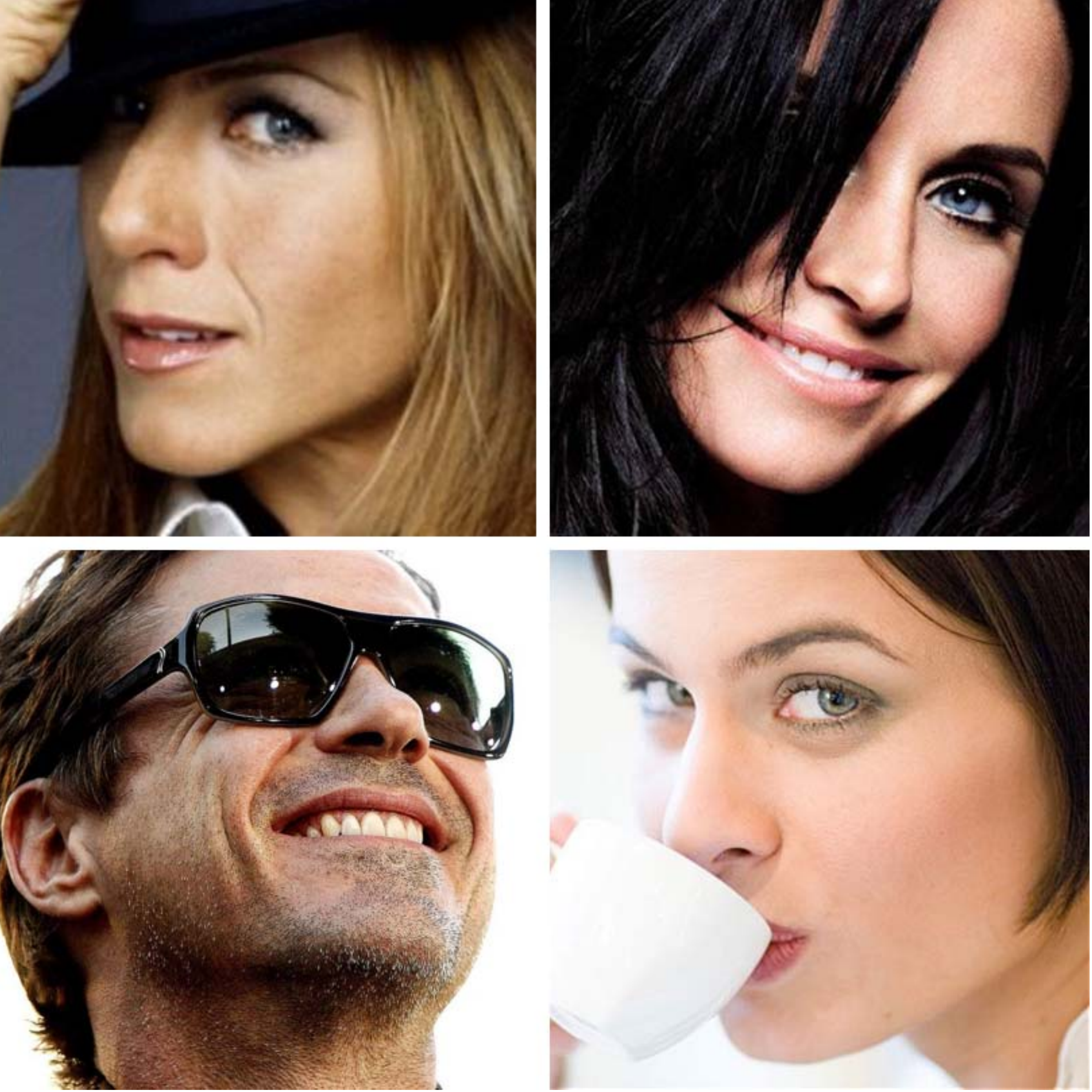}\label{fig_intro_challenge_occlusion}}
\subfigure[Expression]{\includegraphics[width=0.24\textwidth]{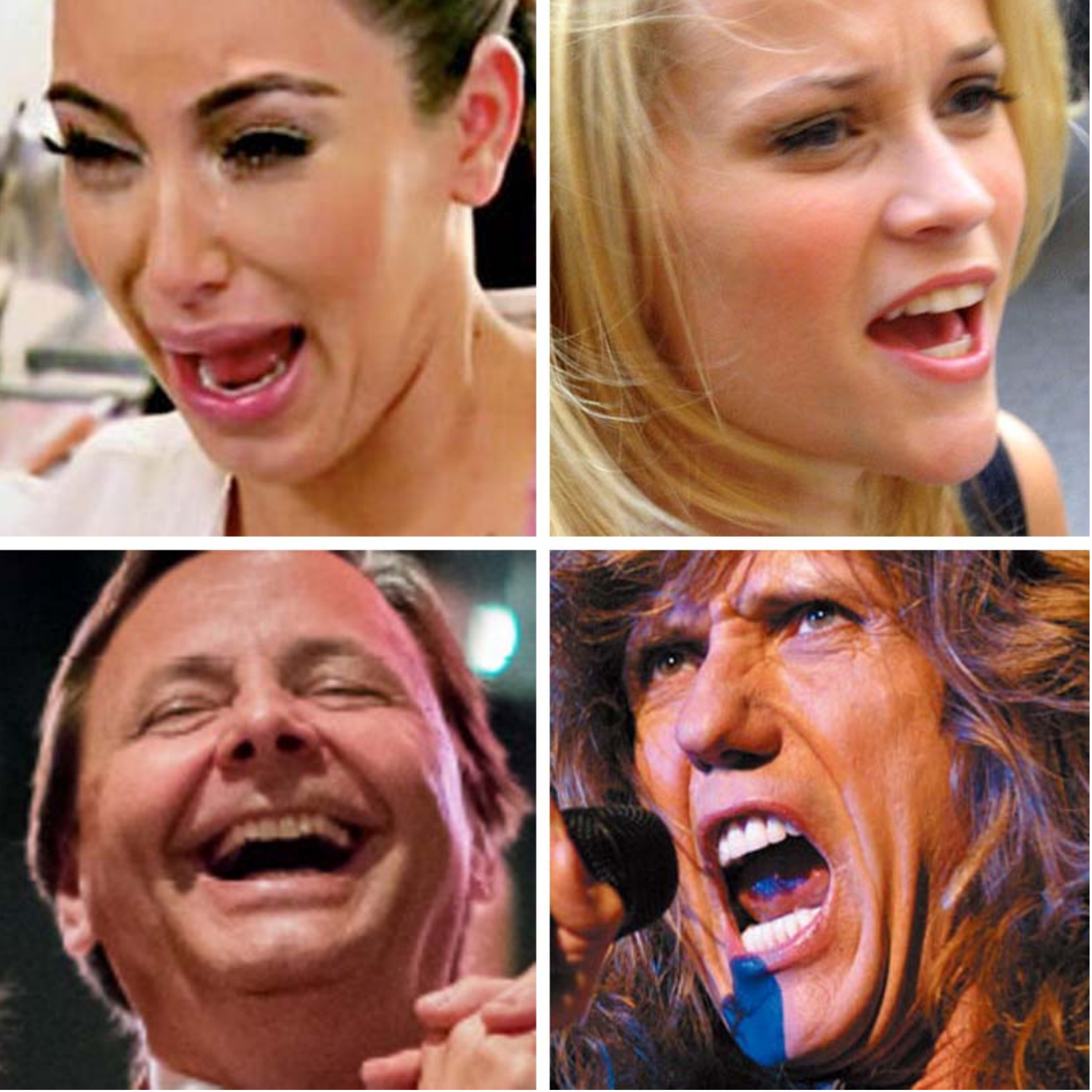}\label{fig_intro_challenge_expression}}
\subfigure[Illumination]{\includegraphics[width=0.24\textwidth]{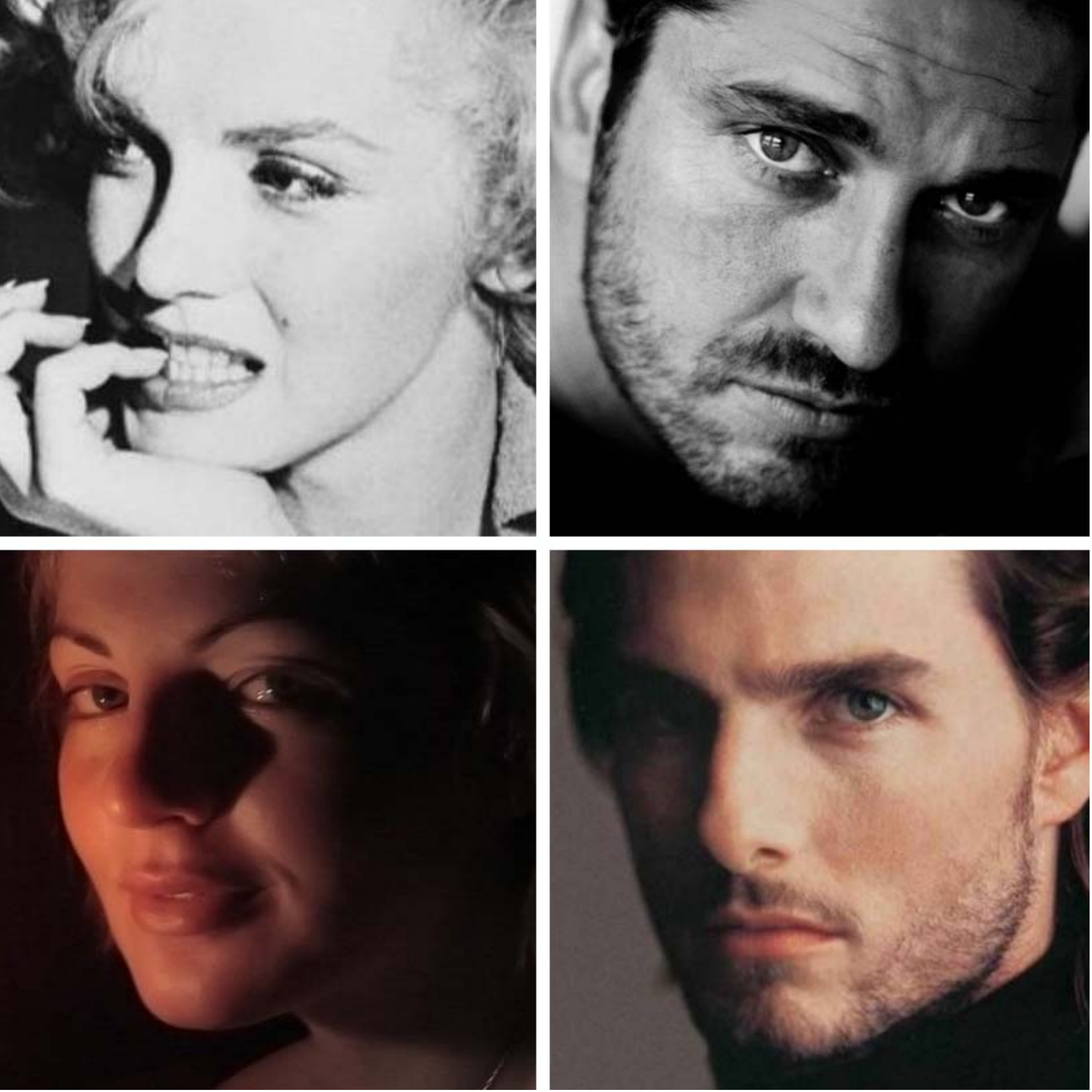}\label{fig_intro_challenge_illumination}}
}
\caption{An illustration of the great challenges of face alignment in the wild (IBUG \cite{sagonas2013300}), from left to right (every two columns): variations in pose, occlusion, expression and illumination.}
\label{fig_intro_challenge}
\end{figure*}
\item \emph{Expression recognition:} The configurations of facial points (typically between 20-60) are reliable indicative of the deformations caused by expressions, and the subsequent analysis will reveal the particular type of expression that may lead to such deformation. Many works \cite{rudovic2010coupled,valstar2012fully,senechal2011combining,bailenson2008real,li2015efficient} follow this idea and use various features extracted from these points for expression recognition.
\end{itemize}
The above-mentioned applications, as well as numerous ones yet to be conceived, urge the need for developing robust and accurate face alignment techniques in real-life scenarios.

Under constrained environments or on less challenging databases, the problem of face alignment has been well addressed, and some algorithms even achieve performance that is close to that of human beings \cite{belhumeur2011localizing,dantone2012real}. Under unconstrained conditions, however, this task is extremely challenging and far from being solved, due to the high degree of facial appearance variability caused either by intrinsic dynamic features of the facial components such as eyes and mouth, or by ambient environment changes. In particular, the following factors have significant influence on facial appearance and the states of local facial features:
\begin{itemize}
\item \emph{Pose:} The appearance of local facial features differ greatly between different camera-object poses (e.g., frontal, profile, upside down), and some facial components such as the one side of the face contour, can even be completely occluded in a profile face.
\item \emph{Occlusion:} For face images captured in unconstrained conditions, occlusion frequently happens and brings great challenges to face alignment. For example, the eyes may be occluded by hair, sunglasses, or myopia glasses with black frames.
\item \emph{Expression:} Some local facial features such as eyes and mouth are sensitive to the change of various expressions. For example, laughing may cause the eyes to close completely, and largely deform the shape of the mouth.
\item \emph{Illumination:} Lighting (varying in spectra, source distribution, and intensity), may significantly change the appearance of the whole face, and make the detailed textures of some facial components missing.
\end{itemize}
These challenges are illustrated in Fig. \ref{fig_intro_challenge} by the IBUG database \cite{sagonas2013300}. An ideal face alignment system should be robust to these facial variations on one hand; while on the other hand, as efficient as possible to satisfy the need of practical applications (e.g., real-time face tracking).

Over the last two decades, numerous techniques have been developed for face alignment with varying degrees of success.  {\c{C}}eliktutan \emph{et al.} \cite{cceliktutan2013comparative} surveyed many traditional methods for face alignment of both 2D and 3D faces, but some recent state-of-the-art methods are not covered. Wang \emph{et al.} \cite{wang2014facial} gave a more comprehensive survey of face alignment methods over the last two decades, but the overall difficulties and challenges in unconstrained environments have not been highlighted. More recently, Yang \emph{et al.} \cite{yang2015empirical} provided an empirical study of recent face alignment methods, aiming to draw some empirical yet useful conclusions and make insightful suggestions for practical applications.

The significant contribution of this paper is to give a comprehensive and critical survey of the ad hoc face alignment methods addressing the difficulties and challenges in unconstrained environments, which we believe would be a useful complement to \cite{cceliktutan2013comparative,wang2014facial,yang2015empirical}. To be self-contained, some traditional methods for face alignment covered in \cite{cceliktutan2013comparative,wang2014facial} are also included. However, contrary to the previous works, we pay special attention to study and summarize the motivation and successful experiences behind the state-of-the-art, expecting to offer some insights into the studies of this field. Furthermore, we organize special discussions on the practical aspects of constructing a face alignment system, including training data augmentation, face preprocessing, shape initialization, accuracy and efficiency tradeoffs. This in our opinion is a very important topic in practice, but is mostly ignored in previous studies. In addition, we show comparative performance statistics of the state of the art, and propose several promising directions for future research.

In the following Section \ref{sec_categorization}, we briefly describe the main idea of face alignment and categorize existing methods into two main categories. Then, the prominent methods within each category are reviewed and analyzed in Section \ref{sec:generative_methods} and \ref{sec:discriminative_methods}. In Section \ref{sec_development}, we investigate some practical aspects of developing of a robust face alignment system. In Section \ref{sec_evaluation}, we discuss a few issues concerning performance evaluation. Finally, we conclude this paper with a discussion of several promising directions for further research in Section \ref{sec_conclusion}.

\section{Overview}\label{sec_categorization}
\begin{table*}[!htb]
\centering
\scriptsize
\caption{Categorization of the popular approaches for face alignment.}
\label{tab_categorization}
\begin{threeparttable}
\begin{tabular}{p{0.32\textwidth}p{0.63\textwidth}}
\Xhline{1pt}
\bf{Approach} & \bf{Representative works}\\
\Xhline{0.5pt}
\bf{Generative methods} & \\
\hspace{0.25em}\emph{Active appearance models (AAMs)} & \\
\hspace{1.5em}Regression-based fitting & Original AAM \cite{cootes2001active}; Boosted Appearance Model \cite{liu2007generic}; Nonlinear discriminative approach \cite{saragih2007nonlinear}; Accurate regression procedures for AMMs \cite{sauer2011accurate}\\
\hspace{1.5em}Gradient descent-based fitting & Project-out inverse compositional (POIC) algorithm \cite{matthews2004active}; Simultaneous inverse compositional (SIC) algorithm \cite{gross2005generic}; Fast AAM \cite{tzimiropoulos2013optimization}; 2.5D AAM \cite{martins2013generative}; Active Orientation Models \cite{tzimiropoulos2014active}\\
\hspace{0.25em}\emph{Part-based\! generative\! deformable\! models} & Original Active Shape Model (ASM) \cite{cootes1995active}; Gauss-Newton deformable part model \cite{tzimiropoulos2014gauss}; Project-out cascaded regression \cite{tzimiropoulos2015project}; Active pictorial structures \cite{antonakos2015active}\\
\bf{Discriminative methods} &\\
\hspace{0.25em}\emph{Constrained local models (CLMs)}\tnote{a}&\\
\hspace{1.5em}PCA shape model & Regularized landmark mean-shift \cite{saragih2011deformable}; Regression voting-based shape model matching \cite{cootes2012robust}; Robust response map fitting \cite{asthana2013robust}; Constrained local neural field \cite{baltrusaitis2013constrained}\\
\hspace{1.5em}Exemplar shape model &  Consensus of exemplar \cite{belhumeur2011localizing}; Exemplar-based graph matching \cite{zhou2013exemplar}; Robust Discriminative Hough Voting \cite{jin2016face}\\
\hspace{1.5em}Other shape models & Gaussian Process Latent Variable Model \cite{huang2007component}; Component-based discriminative search \cite{liang2008face}; Deep face shape model \cite{wu2015discriminative}\\
\hspace{0.25em}\emph{Constrained local regression} &  Boosted regression and graph model \cite{valstar2010facial}; Local evidence aggregation for regression \cite{martinez2013local}; Guided unsupervised learning for model specific models \cite{jaiswal2013guided}\\
\hspace{0.25em}\emph{Deformable part models (DPMs)} & Tree structured part model \cite{zhu2012face}; Structured output SVM \cite{uvrivcavr2012detector}; Optimized part model \cite{yu2013pose}; Regressive Tree Structured Model \cite{hsu2015regressive}\\
\hspace{0.25em}\emph{Ensemble regression-voting} & Conditional regression forests \cite{dantone2012real}; Privileged information-based conditional regression forest \cite{yang2013privileged}; Sieving regression forest votes\cite{yang2013sieving}; Nonparametric context modeling \cite{smith2014nonparametric}\\
\hspace{0.25em}\emph{Cascaded regression} \\
\hspace{1.5em} Two-level boosted regression & Explicit shape regression \cite{cao2012face}; Robust cascaded pose regression \cite{burgos2013robust}; Ensemble of regression trees \cite{kazemi2014one}; Gaussian process regression trees \cite{lee2015face}; \\
\hspace{1.5em} Cascaded linear regression & Supervised descent method \cite{xiong2013supervised}; Multiple hypotheses-based regression \cite{yan2013learn}; Local binary feature \cite{ren2014face}; Incremental face alignment \cite{asthana2014incremental}; Coarse-to-fine shape search \cite{zhu2015face} \\
\hspace{0.25em}\emph{Deep neural networks}\tnote{b} \\
\hspace{1.5em} Deep CNNs & Deep convolutional network cascade \cite{sun2013deep}; Tasks-constrained deep convolutional network \cite{zhang2014facial}; Deep Cascaded Regression\cite{lai2015deep}\\
\hspace{1.5em} Other deep networks & Coarse-to-fine Auto-encoder Networks (CFAN) \cite{zhang2014coarse}; Deep face shape model \cite{wu2015discriminative}\\
\Xhline{1pt}
\end{tabular}
\begin{tablenotes}
    \tiny
    \item[a] Classic Constrained Local Models (CLMs) typically refer to the combination of local detector for each facial point and the parametric Point Distribution Model \cite{cristinacce2006feature,wang2008enforcing,saragih2011deformable}. Here we extend the range of CLMs by including some methods based on other shape models (i.e., exemplar-based model \cite{belhumeur2011localizing}). In particular, we will show that the exemplar-based method \cite{belhumeur2011localizing} can also be interpreted under the conventional CLM framework.
    \item[b] We note that some deep learning-based systems can also be placed in other categories. For instance, some systems are constructed in a cascade manner \cite{zhang2014coarse,lai2015deep,trigeorgis2016mnemonic}, and hence can be naturally categorized as cascaded regression. However, to highlight the increasing important role of deep learning techniques for face alignment, we organize them together for more systematic introduction and summarization.
\end{tablenotes}
\end{threeparttable}
\end{table*}

The problem of face alignment has a long history in computer vision, and a large number of approaches have been proposed to tackle it with varying degrees of success. From an overall perspective, face alignment can be formulated as a problem of searching over a face image for the pre-defined facial points (also called face shape), which typically starts from a coarse initial shape, and proceeds by refining the shape estimate step by step until convergence. During the search process, two different sources of information are typically used: facial appearance and shape information. The latter aims to explicitly model the spatial relations between the locations of facial points to ensure that the estimated facial points can form a valid face shape. Although some methods make no explicit use of the shape information, it is common to combine these two sources of information.

Before describing specific and prominent algorithms, a clear and high-level categorization will help to provide a holistic understanding of the commonality and differences of existing methods in using the appearance and shape information. For this, we follow the basic modeling principles in pattern recognition, and roughly divide existing methods into two categories: \emph{generative} and \emph{discriminative}.
\begin{itemize}
\item \emph{Generative methods:} These methods build generative models for both the face shape and appearance. They typically formulate face alignment as an optimization problem to find the shape and appearance parameters that generate an appearance model instance giving best fit to the test face. Note that the facial appearance can be represented either by the whole (warped) face, or by the local image patches centered at the facial points.
\item \emph{Discriminative methods:} These methods directly infer the target location from the facial appearance. This is typically done by learning independent local detector or regressor for each facial point and employing a global shape model to regularize their predictions, or by directly learning a vectorial regression function to infer the whole face shape, during which the shape constraint is implicitly encoded.
\end{itemize}

Table \ref{tab_categorization} summarizes algorithms and representative works for face alignment, where we further divide the generative methods and discriminative methods into several subcategories. A few methods overlap category boundaries, and are discussed at the end of the section where they are introduced. Below, we discuss the motivation and general approach of each category first, and then, give the review of prominent algorithms within each category, discussing their advantages and disadvantages.

\section{Generative methods}\label{sec:generative_methods}
Typically, faces are modelled as deformable objects which can vary in terms of shape and appearance. Generative methods for face alignment construct parametric models for facial appearance similar to EigenFace \cite{turk1991face}, but differ from EigenFace in that they take into account the deformation of face shape, and build appearance model in a canonical reference frame where the shape variations have been removed. Fitting a generative model aims to find the shape and appearance parameters that can generate a model instance fitting best to the test face.

According to the type of facial representation, generative methods can be further divided into two categories: Active Appearance Models (AAMs) that use the holistic representations, and part-based generative deformable models that use part-based representation.

\subsection{Active appearance models} \label{subsec:AAMs}
Active Appearance Models (AAMs), proposed by Cootes \emph{et al.} \cite{cootes2001active}, are \emph{linear} statistical models of both the shape and the appearance of the deformable object. They are able to generate a variety of instances by a small number of model parameters, and therefore have been widely used in many computer vision tasks, such as face recognition \cite{lanitis1997automatic}, object tracking \cite{stegmann2001object} and medical image analysis \cite{stegmann2003fame}. In the field of face alignment, AAMs are arguably the most well-known family of generative methods that have been extensively studied during the last 15 years \cite{cootes2001active,matthews2004active,gross2005generic,tzimiropoulos2013optimization}.

%In this section, we do not intend to give a comprehensive and detailed survey of AAMs, but instead will focus on recent advances on AAMs that make this classic algorithm generalize well to face alignment \emph{in-the-wild}.

In the following, we first briefly introduce the basic AAM algorithm including AAM modeling and fitting, then  summarize and analysis some recent advances on AAM research, and finally present some discussions about the advantages and disadvantages of AAMs.

\subsubsection{Basic AAM algorithm: modeling and fitting}
In the section, we briefly introduce the basic AAM algorithm: modeling and fitting. Note that we do not intend to give a very comprehensive and detailed overview of the basic AAM algorithm, and refer the reader to recent surveys \cite{gao2010review,wang2014facial} for more details.

\vspace{0.25em}
\noindent \emph{\textbf{AAM modeling.}} An AAM is defined by three components, i.e., \emph{shape model}, \emph{appearance model}, and \emph{motion model}. The shape model, which is coined Point Distribution Model (PDM) \cite{cootes1992active}, is built from a collection of manually annotated facial points $\mathbf{s} = (\mathbf{x}_1^T,...,\mathbf{x}_N^T)^T$ describing the face shape, where $\mathbf{x}_i = (x_i,y_i)$ is the 2-D location of the $i$th point. To learn the shape model, the training face shapes are normalized with respect to a global similarity transform (typically using Procrustes Analysis \cite{gower1975generalized})
and Principal Component Analysis (PCA) is applied to obtained a set of linear shape bases. The shape model can be
mathematically expressed as:
\begin{equation}\label{eq:PCA_shape_model}
  \mathbf{s}(\mathbf{p}) = \mathbf{s_0} + \mathbf{S}\mathbf{p},
\end{equation}
where $\mathbf{s}_0 \in \mathcal{R}^{\{2N,1\}}$ is the mean shape, $\mathbf{S} \in \mathcal{R}^{\{2N,n\}}$ and $\mathbf{p} \in \mathcal{R}^{n}$ is the shape eigenvectors and parameters. Furthermore, this shape model need to be composed with a 2D global similarity transform, in order to position a particular shape model instance arbitrarily on the image frame. For this, using the re-orthonormalization procedure described in \cite{matthews2004active}, the final expression for the shape model can be compactly written using \ref{eq:PCA_shape_model} by appending $\mathbf{S}$ with 4 similarity eigenvectors.

The appearance model is obtained by warping the training faces onto a common reference frame (typically defined by the mean shape), and applying PCA onto the warped appearances. Mathematically, the texture model is defined as follows:
\begin{equation}\label{eq:appearance_model}
  \mathbf{A}(\mathbf{c}) = \mathbf{a_0} + \mathbf{A}\mathbf{c},
\end{equation}
where $\mathbf{a}_0 \in \mathcal{R}^{\{F,1\}}$ is the mean appearance, $\mathbf{A}\in \mathcal{R}^{\{F,m\}}$ and $\mathbf{c} \in \mathcal{R}^{\{m,1\}}$ is the appearance eigenvectors and parameters respectively.

To produce the shape-free textures, the motion model plays a role as a bridge between the image frame and the canonical reference frame. Typically, it is a warp function $\mathcal{W}$ that defines how, given a shape, the image should be warped into a canonical reference frame. Popular motion models include piece-wise affine warp \cite{matthews2004active,tzimiropoulos2013optimization} and Thin-Plate Splines warp \cite{baker2001equivalence}.

\vspace{0.25em}
\noindent \emph{\textbf{AAM fitting.}} Given an test image $\mathbf{I}$, AAM fitting aims to find the optimal parameters $\mathbf{p}$ and $\mathbf{c}$ so that the synthesized appearance model instance gives best fit to the test image in the reference frame. Formally, let $\mathbf{I}[\mathbf{p}] = \mathbf{I}(\mathcal{W}(\mathbf{p}))$ denote the vectorized version of the warped test image, then AAM fitting can be formulated as the following optimization problem,
\begin{equation}\label{eq:AAM_objective}
\mathrm{arg} \mathop{\mathrm{min}}_{\mathbf{p},\mathbf{c}} ||\mathbf{I}[\mathbf{p}] - \mathbf{a}_0 - \mathbf{A}\mathbf{c}||^2.
\end{equation}
Solving \ref{eq:AAM_objective} is an iterative process that at each iteration an update of the current model parameters is estimated. In general, there are two main approaches for AAM fitting.

The first approach is to assume a \emph{fixed} relationship between the residual image and the model parameter increments, and learn it via \emph{regression}. For example, in the original AAM paper \cite{cootes1998active}, this relationship is assumed linear and learned by linear regression, while  in \cite{saragih2007nonlinear} a nonlinear repressor is learned via boosting. However, because the basic assumption that the regression functions are $\emph{fixed}$ is incorrect \cite{matthews2004active}, the regression-based fitting strategies are efficient but approximate.

The second approach is to employ a standard gradient descend algorithm. But unfortunately, standard gradient descend algorithms are inefficient when applied to AAMs fitting. Matthews \emph{et al.} \cite{matthews2004active} addressed this problem with a so-called project-out inverse compositional algorithm (POIC) algorithm, which decouples shape from appearance by projecting out appearance variation, and estimates the warp update in the model coordinate frame and then compose it inversely to the current warp. Although POIC is extremely fast, it is also known to have a small convergence radius, i.e., convergence is especially bad when training and testing images differ strongly \cite{gross2005generic}. Different from POIC that projects out the appearance variations, the simultaneous inverse compositional (SIC) algorithm \cite{gross2003lucas} optimizes the shape parameter and appearance parameter simultaneously under the inverse compositional framework. SIC has been shown to be more robust than POIC for generic AAM fitting, but the computational cost is much higher \cite{gross2005generic}. Besides SIC, another accurate AAM fitting algorithm is the Alternating Inverse Compositional (AIC) algorithm \cite{papandreou2008adaptive}, which solves two separate minimization problems, one for the shape and one for the appearance optimal parameters, in an alternating fashion.

\subsubsection{Recent advances on AAMs}
Recently, some extensions and improvements of AAMs have been proposed to make this classic algorithm better adapted to the task of face alignment \emph{in-the-wild}. In general, recent advances on AAMs mainly focus on three aspects: (1) unconstrained training data \cite{tzimiropoulos2013optimization}, (2) robust image representations \cite{tzimiropoulos2012generic,antonakos2014hog} and (3) robust and fast fitting strategies \cite{tzimiropoulos2012generic,tzimiropoulos2013optimization}.

\begin{figure}[!tb]
\centering
\includegraphics[width=0.38\textwidth]{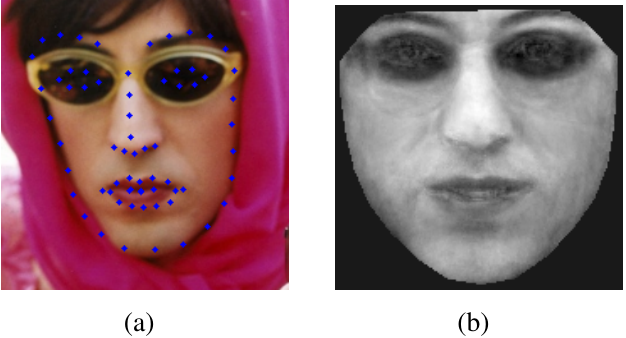}
\caption{(a) A face image from the test set of LFPW \cite{belhumeur2011localizing}, with facial points detected by the Fast-SIC algorithm proposed in \cite{tzimiropoulos2013optimization}. (b)  Reconstruction of the image from the appearance subspace. (Fig. 2 in \cite{tzimiropoulos2013optimization})}
\label{fig:AAM_in-the-wild}
\end{figure}

\vspace{0.25em}
\noindent \emph{\textbf{Unconstrained training data.}} Although some AAM fitting algorithms (e.g., the Simultaneous Inverse Compositional (SIC) algorithm \cite{gross2003lucas}) are known to perform well on constrained face databases, their performance has not been assessed on in-the-wild databases until recently. Tzimiropoulos \emph{et al.} \cite{tzimiropoulos2013optimization} showed that, when trained in-the-wild, AAMs perform notably well and in some cases comparably with current state-of-the-art methods, without using sophisticated shape priors, robust features or robust norms. Fig. \ref{fig:AAM_in-the-wild} shows a fitting case and the reconstruction of the image, produced by AAM built in the wild \cite{tzimiropoulos2013optimization}.

\vspace{0.25em}
\noindent \emph{\textbf{Robust image representations.}} Typically, AAMs use the pixel-based image representation that is sensitive to global lighting \cite{cootes1998active,matthews2004active,tzimiropoulos2013optimization}, and a natural way to improve the robustness of AAMs is to use the feature-based representation. In general, robust image features, such as HOG \cite{dalal2005histograms}, SIFT \cite{lowe2004distinctive} or SURF \cite{bay2008speeded} that describe distinctive and important image characteristics, can generalize better to unseen images. Some recent works have confirmed the robustness of the appearance model built upon feature-based representation \cite{tzimiropoulos2012generic,tzimiropoulos2014active,antonakos2015feature}.

\vspace{0.25em}
\noindent \emph{\textbf{Robust and fast fitting strategies.}} It is widely acknowledged that the Project-out Inverse Compositional (POIC) algorithm is fast but has a small convergence radius, while the Simultaneous Inverse Compositional (SIC) algorithm is accurate but very slow. Due to this, some recent advances on AAMs have focused on robust and fast fitting algorithms. For example, it was found in \cite{tzimiropoulos2012generic,antonakos2015feature} that the Alternating Inverse Compositional (AIC) algorithm \cite{papandreou2008adaptive} performs well for generic AAM fitting. Although AIC is slower than the project-out algorithm, it is still very fast probably allowing a real-time implementation. Furthermore, by using a standard results from optimization theory, Tzimiropoulos \emph{et al.} \cite{tzimiropoulos2013optimization} dramatically reduced the dominant cost for both SIC and the standard Lukas-Kanade algorithm, making both algorithms very attractive speed-wise for practical AAM systems.

\subsubsection{Discussion} We have described the basic AAM algorithm and recent advances on AAMs. Despite the popularity and success, AMMs have been traditionally criticized for the limited representational power of their holistic representation, especially when used in wild conditions. However, recent works on AAMs \cite{tzimiropoulos2012generic,antonakos2014hog,lucey2013fourier} suggest that this limitation might have been over-stressed in the literature and that AAMs can produce highly accurate results if appropriate training data \cite{tzimiropoulos2013optimization}, image representations \cite{tzimiropoulos2012generic,antonakos2014hog} and fitting strategies \cite{tzimiropoulos2012generic,tzimiropoulos2013optimization} are employed.

Despite this, AAMs are still considered to have the following drawbacks: (1) Since the holistic appearance model is used, partial occlusions cannot be easily handled. (2) For the appearance model built \emph{in-the-wild}, the dimension of appearance parameter is very high, which makes AAMs difficult to optimize and likely to converge to undesirable local minima. One possible way to overcome these drawbacks is to use part-based representations, due to the observation that local features are generally not as sensitive as global features to lighting and occlusion. In the following section, we turn to part-based generative methods.

\subsection{Part-based generative deformable models}
Part-based generative methods build generative appearance models for facial parts, typically with a shape model to govern the deformations of the face shapes. In this paper, we do not distinguish the specific form of the shape model, and refer to all part-based generative methods collectively as \emph{part-based generative deformable models}.

In general, there are two approaches to construct generative part models. The first is to construct individual appearance model for each facial part. A notable example is the well-known original Active Shape Models \cite{cootes1992active,cootes1995active} that combine the generative appearance model for each facial part and the Point Distribution Model for global shapes. However, we note that a more natural and popular way is to model individual facial part is the \emph{discriminatively} trained local detector \cite{cristinacce2007boosted,saragih2011deformable,cootes2012robust,asthana2013robust}, as adopted by a very successful family of methods coined Constrained Local Models (CLMs) \cite{saragih2011deformable,asthana2013robust}. Actually, ASMs can be regarded as the predecessors of CLMs, because they are similar in both the models and the fitting process. Therefore, we refer the reader to Section \ref{subsubsec:CLMs} for more details about ASMs under the CLM framework.

The second approach is to construct generative models for all facial parts simultaneously. For example, one can concatenate all facial parts (image patches) to form a part-based representation for the whole face, and then build generative appearance model for it. The Gauss-Newton Deformable Part Model (GN-DPM) \cite{tzimiropoulos2014gauss} has explored this idea, and build linear statistical model for both the concatenated facial parts and the shape using PCA. Benefiting from the part-based representation, the motion model of GN-DPM degenerates to similarity transformation, rather than the affine warp of AAMs. In the fitting phase, GN-DPM formulate and solve the non-linear least squares optimization problem similar to AAMs \cite{matthews2004active,tzimiropoulos2013optimization}. The part-based appearance model along with a global shape model is optimized by the fast SIC algorithm \cite{tzimiropoulos2013optimization} in a Gauss-Newton fashion. Extensive experiments on wild face databases \cite{belhumeur2013localizing,le2012interactive,zhu2012face} demonstrate that the part-based GN-DPM outperforms AAMs by a large margin.

While GN-DPM employs the inverse compositional fitting algorithm, Tzimiropoulos \emph{et al.} \cite{tzimiropoulos2015project} consider the forward algorithm for the non-linear least square optimization problem akin to that of GN-DPM. Although analytic gradient decent method is employed in \cite{tzimiropoulos2015project}, it is only used for the derivation of the learning and fitting basis of the proposed Project-Out Cascaded Regression (PO-CR) method. In particular, PO-CR learns from data a sequence of averaged Jacobians and descent directions via regression in a subspace orthogonal to the facial appearance variation. Apart from the PCA-based appearance model in GN-DPM and PO-CR, Antonakos \emph{et al.} \cite{antonakos2015active} propose to model the appearance of facial parts using multiple pairwise distributions based on the edges of a graph (GMRF), and show that this outperforms the commonly used PCA model under an inverse Gauss-Newton optimization framework.

Compared to AAMs, the part-based generative deformable models mainly have the advantages from part-based representation, i.e., more robust to global lighting and occlusion in wild conditions. As shown in \cite{tzimiropoulos2014gauss}, part-based generative models may have the same representational power of AAMs, but are easier to optimize. That is, when the initial shape is far from the ground truth, part-based generative deformable models are more likely to get converged to a good solution, although the formulation is non-convex by nature.

\subsection{Summary and discussion}
We have reviewed generative methods for face alignment in two categories, i.e., Active Appearance Models (AAMs) that use the holistic representation and the part-based generative deformable models that use the part-based representation. In general, the fitting result of a generative appearance model to a test image typically depends on two factors: (1) the representational power of the model, and (2) the difficulty in optimizing the model. As investigated in \cite{tzimiropoulos2014gauss}, when trained in-the-wild, both AAMs and part-based generative deformable models can reconstruct the appearance of an unseen image well, but the part-based generative deformable models are considered to be easier to optimize than AAMs. Furthermore, recent results show that if unconstrained training data \cite{tzimiropoulos2013optimization}, robust image representations \cite{tzimiropoulos2012generic,antonakos2014hog} and appropriate fitting strategies \cite{tzimiropoulos2012generic,tzimiropoulos2013optimization,tzimiropoulos2014gauss,tzimiropoulos2015project} are employed, generative methods can produce a very high degree of fitting accuracy for face alignment \emph{in-the-wild}. These results suggest that the limitations of generative methods, especially the AAMs, might have been over-stressed in the literature. In addition, generative methods typically have the advantage of requiring fewer training examples than the discriminative methods to perform well \cite{antonakos2015active}.

However, with recent development of unconstrained facial databases with an abundance of annotated facial data captured, the discriminative methods, which are capable of effectively leveraging large bodies of training data, are arguably now playing a more and more prominent role in face alignment \emph{in-the-wild}. Next, we will turn to discriminative methods.

\section{Discriminative methods}\label{sec:discriminative_methods}
\begin{table*}[!htb]
\centering
\scriptsize
\caption{Overview of the six classes of discriminative methods in our taxonomy.}
\label{tab_discriminative_description}
\begin{threeparttable}
\begin{tabular}{p{0.21\textwidth}p{0.22\textwidth}p{0.16\textwidth}p{0.33\textwidth}}
\Xhline{1pt}
 & \bf{Appearance model} & \bf{Shape model} & \bf{Highlights of the method}\\
\Xhline{1pt}
\emph{Constrained local models} & Independently trained local detector that computes a pseudo probability of the target point occurring at a particular position. & Point Distribution Moldel; Exemplar model, etc\tnote{a}. &  The local detectors are first correlated with the image to yield a filter response for each facial point, and then shape optimization is performed over these filter responses.\\ \\
\emph{Constrained local regression} & Independently trained local regressor that predicts a distance vector relating to a patch location. & Markov Random Fields to model the relations between relative positions of pairs of points. & Graph model is used to constrain the search space of local regressors by exploiting the constellations that facial points can form.\\ \\
\emph{Deformable part models} & Part-based appearance model that computes the appearance evidence for placing a template for a facial part. & Tree-structured models that are easier to optimize than dense graph structures. & All parameters of the appearance model and shape model are discriminatively learned in a max-margin structured prediction framework; efficient dynamic programming algorithms can be used to find globally optimal solutions.\\ \\
\hline
\emph{Ensemble regression-voting} & Image patches to cast votes for all facial points relating to the patch centers;

Local appearance features centered at facial points. & Implicit shape constraint that is naturally encoded into the multi-output function (e.g., regression tree).  & Votes from different regions are ensembled to form a robust prediction for the face shape.\\ \\
\emph{Cascaded regression} & Shape-indexed feature that is related to current shape estimate (e.g., concatenated image patches centered at the facial points). & Implicit shape constraint that is naturally encoded into the regressor in a cascaded learning framework. & Cascaded regression typically starts from an initial shape (e.g., mean shape), and refines the holistic shape through sequentially trained regressors.\\ \\
\emph{Deep neural networks} & Whole face region that is typically used to estimate the whole face shape jointly;

Shape-indexed feature\tnote{b}. & Implicit shape constraint that is encoded into the networks since all facial points are predicted simultaneously. & Deep network is a good choice to model the nonlinear relationship between the facial appearance and the shape update.

Among others, deep CNNs have the capacity to learn highly discriminative features for face alignment.\\
\Xhline{1pt}
\end{tabular}
\begin{tablenotes}
    \tiny
    \item[a] Constrained Local Models (CLMs) typically employ a parametric (PCA-based) shape model \cite{saragih2011deformable}, but we will show that the exemplar-based method \cite{belhumeur2011localizing} can also be derived from the CLM framework. Furthermore, we extend the range of CLMs by including some methods that combine independently local detector and other face shape model \cite{huang2007component,liang2008face,wu2015discriminative}.
    \item[b] Some deep network-based systems follow the cascaded regression framework, and use the shape-indexed feature \cite{zhang2014coarse}.
\end{tablenotes}
\end{threeparttable}
\end{table*}

Discriminative face alignment methods seek to learn a (or a set of) discriminative function that directly maps the facial appearance to the target facial points. In general, there are two main lines of research for discriminative methods. The first line is to follow the ``divide and conquer" strategy by learning discriminative local appearance model (detector or regressor) for each facial point, and a shape model to impose global constraints on these local models. This line can be further subdivided into three classes: (1) \emph{Constrained Local Models (CLMs)} that learn independent local detector for each facial point, with a shape model to regularize the detection responses of these local detectors. (2) \emph{constrained local regression} methods that learn independent local regressor for each point and use a graph model to guide the search of these local regressors, and (3) \emph{deformable part models} that learn the local appearance model and the tree structured shape model jointly in a discriminative framework.

The second line is to directly learn a vectorial regression function to infer the \emph{whole} face shape, during which the shape constraint is implicitly encoded. This line can also be further subdivided into three classes: (1) \emph{ensemble regression-voting} methods that cast votes for all facial points from local regions via regression, and ensemble the votes from different regions to form a robust prediction, (2) \emph{cascaded regression} methods that learn a vectorial regression function in a cascade manner to estimate the face shape stage-by-stage, and (3) \emph{deep neural networks} that employ deep convolutional networks \cite{sun2013deep,zhang2014facial} or auto-encoder networks \cite{zhang2014coarse} to model the nonlinear relationship between the facial appearance and the shape update.

Table \ref{tab_discriminative_description} gives a overview of the six classes of discriminative methods in our taxonomy, where the appearance model, shape model and highlights of them are listed respectively to show the differences and relations between them.

\subsection{Constrained local models} \label{subsubsec:CLMs}
Constrained Local Models (CLMs), which can date back to the seminal work of Active Shape Model (ASM) \cite{cootes1995active}, are a relatively mature approach for face alignment \cite{cristinacce2006feature,saragih2011deformable,cootes2012robust,asthana2013robust,baltrusaitis2013constrained}. In the training phase, CLMs learn independent local detector for each facial point, and a prior shape model to characterize the deformation of face shapes. In testing, face alignment is typically formulated as an optimization problem to find the best fit of the shape model to the test image. We classify CLMs as the discriminative methods because of the discriminative nature of usual local detectors.

While the seminal work of \cite{saragih2011deformable} unifies various CLM approaches in a probabilistic framework, it only focuses on the CLMs using the Point Distribution Model (PDM). However, we note that some methods using other shape model (i.e., the exemplar shape model \cite{belhumeur2011localizing}) are also close to \cite{saragih2011deformable} in methodology. Hence, in this paper we refer to those methods combining independent local detector and any kind of shape model collectively as \emph{Constrained Local Models}\footnote{A disadvantage of our extended definition for CLMs is that, the classical Deformable Part Models (DPMs) with the tree structured shape model \cite{yang2013articulated,zhu2012face} will also be covered by our definition of CLMs, which however are traditionally treated as an independent approach relative to others. In this paper, we will still treat DPMs as an independent group of methods in face alignment, and describe them in Section \ref{subsec:DPMs} }.

In the following, we will first briefly introduce the basic Point Distribution Model (PDM) based CLM algorithm including modeling and fitting, then summarize and analysis recent advances on CLMs in handling unconstrained challenges. In particular, we will show that exemplar-based method \cite{belhumeur2011localizing} can also be interpreted under the conventional CLM framework \cite{saragih2011deformable}. Finally, we discuss the advantages and disadvantages of CLMs.

\subsubsection{Basic CLM algorithm: modeling and fitting}
In this section, we will briefly describe the basic CLM algorithm building upon the Point Distribution Model (PDM), which has two procedures: modeling and fitting.

\vspace{0.25em}
\noindent \emph{\textbf{CLM modeling.}} A CLM consists of two important components: \emph{local detector} for each facial point, and the \emph{shape model} that captures the deformations of valid face shapes. The task of local detector is to compute a pseudo probability (likelihood) that the target point occurs at a particular position. Existing local detectors can be broadly categorized into three groups.
\begin{itemize}
\item \emph{Generative approach:} Generative approaches can be use to model local image patches centered at the annotated facial points. For example, \cite{cootes1992active,cootes1993active} assume that the local appearance is multivariate Gaussian distributed, and use the Mahalanobis distance as the fitting response for a new image patch.
\item \emph{Discriminative classifier:} Discriminative classifier-based approach learns a binary classifier for each point with annotated image patches to discriminate whether the target point is aligned or not when testing. To cast various CLM fitting strategies in a unified probabilistic framework, the output of these classifiers are typically transformed into pseudo probabilities. Different types of classifiers have been exploited in literature, e.g., logistic regression \cite{saragih2011deformable}, SVM \cite{belhumeur2011localizing,asthana2013robust}, and Local Neural Field (LNF) \cite{baltrusaitis2013constrained}.
\item \emph{Regression-voting approach:} The regression-voting approach casts votes for the target point from a nearby region, then compute the pseudo probabilities by accumulating votes from different regions \cite{cristinacce2007boosted,cootes2012robust}. The regression-voting approach has the potential to be more efficient since a locally exhaustive search is avoided.
\end{itemize}

Due to the local patch support and large variations in training, the local detectors are typically imperfect, and the correct location will not always be at the location with the highest detection response. To address this drawback, a global shape model is typically employed to regularize the detection of these local detectors. For this, conventional CLMs use the Point Distribution Model (PDM) that simply models the normalized face shapes as multivariate Gaussian and approximates them using PCA (see Equation \ref{eq:PCA_shape_model}).

\vspace{0.25em}
\noindent \emph{\textbf{CLM fitting.}} Overall, give an image $\mathbf{I}$, the goal of PDM-based CLMs is to find the optimal shape parameter $\mathbf{p}$ that maximizes the probability of its points corresponding to consistent locations of the facial features. By assuming that the local search of each facial point is conditionally independent, the fitting objective of PDM-based CLMs can be written as:
\begin{equation}\label{eq:PCA_CLM_goal}
\begin{split}
\mathbf{p^*} &= \mathrm{\mathop{arg \, max}_{\mathbf{p}}} \; p(\mathbf{p}|\{l_i\!=\!1\}_{i=1}^N, \mathbf{I}) \\
&= \mathrm{\mathop{arg \, max}_{\mathbf{p}}} \; p(\mathbf{p})\prod_{i=1}^N p(l_i\!=\!1|\mathbf{x}_i(\mathbf{p}),\mathbf{I}),
\end{split}
\end{equation}
where $\mathbf{x}_i(\mathbf{p})$ is the location of the $i$th point generated by the shape model, $l_i \in \{1,-1\}$ is a discrete random variable denoting whether the $i$th facial point is aligned or not, and $p(\mathbf{p})$ is the prior distribution of $\mathbf{p}$ that can be estimated from the training data.

\begin{figure}[!tb]
\centering
\includegraphics[width=0.48\textwidth]{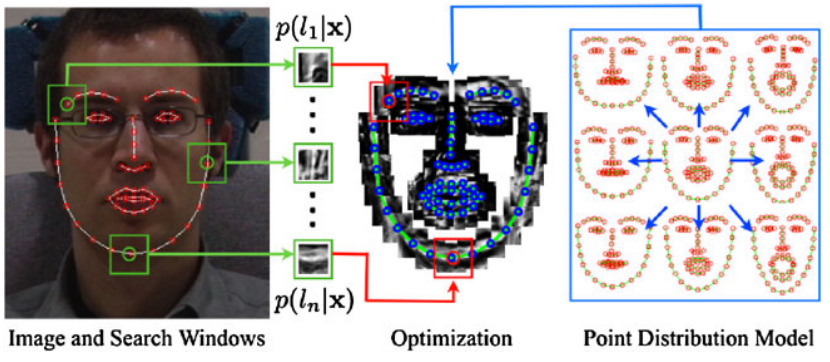}
\caption{ Illustration of PDM-based CLM fitting and its two components: (1) an exhaustive local search for feature locations to get the response maps $\{p(l_i = 1|\mathbf{x}, \mathbf{I})\}_{i=1}^N$ and (2) an optimization strategy to maximize the responses of the PDM constrained facial points. (Fig. 2 in \cite{saragih2011deformable})}
\label{fig:PDM_CLM_fitting}
\end{figure}

CLM fitting based on \ref{eq:PCA_CLM_goal} is an iterative process (see Fig. \ref{fig:PDM_CLM_fitting}) that entails (1) convolving the local detectors with the image to generate response maps, and (2) performing a global shape optimization procedure over these response maps. To make optimization efficient and numerically stable, a common choice of existing optimization strategies is to replace the true response maps with some approximate forms and then perform Guass-Newton optimization over them instead of the original response maps.

\begin{table}[!htb]
  \centering
  \footnotesize
  \caption{Different approximation strategies of response map.}
  \label{tab_responsemap_approximation}
  \begin{tabular}{p{0.37\textwidth}p{0.35\textwidth}}
  \Xhline{1pt}
  & Approximation of response map\\
  \Xhline{0.5pt}
  Isotropic Gaussian estimator \cite{cootes1995active} & $\mathcal{N}(\mathbf{x}_i(\mathbf{p});\bm{\mu}_i,\sigma_i^2\mathbf{I}^{(e)})$\\
  Anisotropic Gaussian estimator \cite{wang2008enforcing} & $\mathcal{N}(\mathbf{x}_i(\mathbf{p});\bm{\mu}_i,\bm{\Sigma}_i)$\\
  Gaussian mixture model \cite{gu2008generative} & $\sum_{k=1}^{K_i}\pi_{ik}\mathcal{N}(\mathbf{x}_i(\mathbf{p});\bm{\mu}_{ik},\bm{\Sigma}_{ik})$\\
  Gaussian kernel estimation \cite{saragih2011deformable} & $\sum_{\mathbf{y}_j \in \bm{\Psi}_{\mathbf{x}_i}}\pi_{\mathbf{y}_j}\mathcal{N}(\mathbf{x}_i(\mathbf{p});\mathbf{y}_j,\rho^2\mathbf{I}^{(e)})$\\
  \Xhline{1pt}
  \end{tabular}
\end{table}

The seminal framework of \cite{saragih2011deformable} unifies various approximation strategies for the true response maps. As listed in Table \ref{tab_responsemap_approximation}, they are (1) the isotropic Gaussian estimators used by original ASMs \cite{cootes1992active,cootes1995active}, where $\bm{\mu}_i$ is the the location of the maximum filter response within the $i$th response map, and $\sigma_i^{-2}$ is the detection confidence over peak response coordinate, (2) a full covariance anisotropic Gaussian estimators used in \cite{wang2008enforcing}, where $\bm{\Sigma}_i$ is the anisotropic covariance matrix of Gaussian distribution, (3) Gaussian mixture model (GMM) used in \cite{gu2008generative}, where $K_i$ denotes the number of modes and $\{\pi_{ik}\}_{k=1}^{K_i}$ are the mixing coefficients for the GMM of the $i$th point, and (4) a homoscedastic isotropic Gaussian kernel estimation (KDE) used by \cite{saragih2011deformable}, where $\pi_{\mathbf{y}_j} = p(l_i=1|\mathbf{y}_j,\mathbf{I})$ denotes the likelihood that the $i$th point is aligned at location $\mathbf{y}_j$, and $\rho^2$ denotes the variance of the noise on facial point locations, $\mathbf{I}^{(e)}$ is the identity matrix. Among them, the nonparametric Gaussian kernel estimation (KDE) method \cite{saragih2011deformable} is considered to achieve a good tradeoff between representation power and the computational complexity. This method is known as Regularized Landmark Mean-Shift (RLMS) fitting, as the resulting update equations based on this nonparametric approximation are reminiscent of the well known mean-shift \cite{fukunaga1975estimation} over the facial point but with regularization imposed by the Point Distribution Model.

Due to its effectiveness and efficiency, the RLMS method \cite{saragih2011deformable} has been extensively investigated. For example, Baltruvsaitis \emph{et al.} \cite{baltruvsaitis20123d} explored the information of depth images, and extend the RLMS \cite{saragih2011deformable} algorithm to a 3D vision. Unlike aforementioned approximations to response maps, \cite{asthana2013robust} proposes a novel discriminative regression based approach to directly estimate the parameter update, and results in significant performance improvement.

\subsubsection{Recent advances on CLMs} \label{subsec:recent_advances_CLMs}
Recently, some improvements of the conventional CLMs have been proposed to better handle various challenges in-the-wild. In general, recent advances on CLMs mainly focus on three aspects: (1) better local detectors, (2) discriminative fitting, and (3) other shape models.

\vspace{0.25em}
\noindent \emph{\textbf{Better local detectors.}} Conventional CLMs typically use logistic regression \cite{saragih2011deformable} or SVM \cite{belhumeur2011localizing,asthana2013robust} to train local detector, which however is plagued by the problem of ambiguity, especially on the wild databases. To mitigate this ambiguity, some advanced local detectors have been proposed, such as the Minimum Output Sum of Squared Errors (MOSSE) filters \cite{martins2014non} and the Local Neural Field (LNF) patch expert, which are able to capture more complex information and exploit spatial relationships between pixels, and hence can achieve better detection results.

\vspace{0.25em}
\noindent \emph{\textbf{Discriminative fitting.}} It is widely acknowledged that the formulation based on CLMs is non-convex, and in general prone to local minima. As an alternative, Asthana \emph{et al.} \cite{asthana2013robust} proposed a novel Discriminative Response Map Fitting (DRMF) method for the CLM fitting that outperforms the RLMS fitting method \cite{saragih2011deformable} in wild databases. We conjecture that the robustness of DRMF mainly stems from the discriminative training process, which can effectively leverage large bodies of training data.

\vspace{0.25em}
\noindent \emph{\textbf{Other shape models.}} One problem with the Point Distribution Model (PDM) is that its the model flexibility is heuristically determined by PCA dimension. To overcome this drawback, some other shape models are proposed to combine with the local detectors for face alignment \cite{huang2007component,belhumeur2011localizing,wu2015discriminative}. In particular, we will show that the exemplar-based method \cite{belhumeur2011localizing} can be derived and well interpreted under the conventional CLM framework \cite{saragih2011deformable}.

The exemplar-based method \cite{belhumeur2011localizing} assumes that the face shape $\mathbf{s} = (\mathbf{x}_1,...,\mathbf{x}_N)^T$ in the test image is generated by one of the transformed exemplar shapes (global models). Let $\mathbf{s}_{k,t}$ ($k=1,...,D$) denote locations of all facial points in the $k$th of the $D$ exemplars that transformed by some similarity transformation $t$, and let $\mathbf{x}_{i,k,t}$ denote location of the $i$th facial point of the transformed exemplar $\mathbf{s}_{k,t}$. By assuming that conditioned on the global model $\mathbf{s}_{k,t}$, the location of each facial point $\mathbf{x}_i$ is conditionally independent of one another, the exemplar-based shape model $p(\mathbf{s})$ can be written as follows:
\begin{equation}\label{eq:exemplar_shape_model}
\begin{split}
p(\mathbf{s}) & = \sum_{k=1}^D \int_{t \in{T}}p(\mathbf{s},\mathbf{s}_{k,t})dt \\
& = \sum_{k=1}^D \int_{t\in{T}} \prod_{i=1}^N p(\mathbf{x}_i|\mathbf{x}_{i,k,t})p(\mathbf{s}_{k,t})dt,
\end{split}
\end{equation}
where $p(\mathbf{x}_i|\mathbf{x}_{i,k,t})$ is modeled as a Gaussian distribution centered at $\mathbf{x}_{i,k,t}$, and the prior of the global model $p(\mathbf{s}_{k,t})$ is assumed as an uniform distribution. Then, by replacing the shape model $p(\mathbf{p})$ in conventional CLM framework \ref{eq:PCA_CLM_goal} with above exemplar-based model $p(\mathbf{s})$, we derive the objective function of \cite{belhumeur2011localizing} (difference in notations) as follows:
\begin{equation}\label{eq:exemplar_goal}
\mathbf{s}^* \! =\! \mathrm{\mathop{arg \, max}_{\mathbf{s}}} \, \sum_{k=1}^D \! \int_{t\in{T}} \prod_{i=1}^N p(\mathbf{x}_i|\mathbf{x}_{i,k,t})p(l_i\!=\!1|\mathbf{x}_i,\mathbf{I})dt.
\end{equation}
This function is optimized by employing RANSAC to sample global models. Due to the use of RANSAC, the exemplar-based method \cite{belhumeur2011localizing} has two advantages over conventional CLMs: (1) independent of shape initialization, and (2) robust to partial occlusion, and achieves excellent performance on the wild LFPW database \cite{belhumeur2011localizing}.

The global models in \cite{belhumeur2011localizing} are scored and selected by the global likelihood, i.e., multiplying the detection response of each local detector. However, as pointed by Jin \emph{et al.} \cite{jin2016face}, this global likelihood score function ignores the difference between local detectors, while in fact, an eye detector is typically more reliable than a chin detector. In \cite{jin2016face}, a discriminatively trained score function is proposed to evaluate the goodness of a global model, which weighs the importance of different local detectors. Furthermore, an efficient pipeline was proposed in \cite{jin2016face} to alleviate the effect of inaccurate anchor points for generating global models.

\subsubsection{Discussion} We have reviewed the basic CLM algorithm and recent advances. In general, CLMs are considered to be more robust to partial occlusion and global lighting than the holistic approaches (e.g., AAMs) \cite{saragih2011deformable}, due to their part-based modeling. However, the local detectors of CLMs are imperfect and have been shown to result in detection ambiguities in testing. Furthermore, since the global shape optimization is performed on the response maps, the detection ambiguities may lead to performance bottleneck, when facing various challenges in unconstrained conditions.

Another disadvantage of CLMs is that they perform an expensive locally exhaustive search for each facial point. One way to reduce the computational cost is to use a displacement expert (local regressor, i.e., estimate the relative position of the target point with respect to the given patch. We will turn to this topic in the next section.

\subsection{Constrained local regression}
Besides the CLMs, another local model-based approach is to train independent local \emph{regressor} for each point, and employ a global shape model to restrict the search of these local regressors to anthropomorphically consistent regions \cite{valstar2010facial,martinez2013local}. Since this idea is similar to CLMs, we refer to this approach as \emph{constrained local regression}.

A representative work of this group is the Boosted Regression coupled with Markov Netwroks \cite{valstar2010facial} (BoRMaN) method, which iteratively uses Support Vector Regressoin (SVR) to provide an initial prediction for all points, and then applies the Markov Network to ensure that the new locations sampled to apply the local regressors are from correct point constellations. BoRMaN let each node in the graph associated to a spatial relation between two points and define pairwise relations between nodes, which allows a representation that is invariant to in-plane rotations, scale changes and translations. Essentially, BoRMaN performs an iterative sequential refinement of the estimate, where the previous target estimate becomes the test location at the next iteration. Martinez \emph{et al.} \cite{martinez2013local} argue that this sequential estimation approach has a series of drawbacks, for example, sensitive to the starting point and any errors in the estimation process. To improve the robustness of BoRMaN, \cite{martinez2013local} propose to detect the target location by aggregating the estimates obtained from stochastically selected local appearance information into a single robust prediction, and refer to their algorithm as Local Evidence Aggregated Regression (LEAR).

The main advantage of constrained local regression approach is that combing local regressors with MRF may drastically reduce the time needed to search for point location, while its disadvantages are: (1) similar to CLMs, its performance is limited by the detection ambiguities of the independently trained local regressors, and (2) globally optimizing MRF is intractable. An alternative choice to the graph-based MRF are the tree-structured models, which are also effective to capture global elastic deformation, but easier to optimize than MRF.

\subsection{Deformable part models} \label{subsec:DPMs}
The tree-structured models are a natural and effective choice to model deformable objects \cite{yang2013articulated,zhu2012face}, which benefit from the existence of an efficient dynamic programming algorithms \cite{felzenszwalb2005pictorial} for finding globally optimal solutions. Actually, discriminatively trained tree-structured models have been successfully explored in many computer vision tasks, such as object detection \cite{felzenszwalb2010object}, human pose estimation \cite{yang2013articulated}, and recently in face alignment \cite{zhu2012face,uvrivcavr2012detector,hsu2015regressive}. We follow the nomenclature of \cite{felzenszwalb2010object} and refer to them collectively as \emph{deformable part models} (DPMs).

The main challenges of applying tree-structured model for face alignment may lie in the fact that a single tree-structured pictorial structure, perhaps, is insufficient to capture various shape deformations due to viewpoint. This problem is addressed by the seminal work of Zhu \emph{et al.} \cite{zhu2012face}, with a unified framework for face detection, pose estimation and face alignment. They modeled every facial point as a part and used mixtures of trees to capture the global topological changes due to viewpoint; a part will only be visible in certain mixtures/views. Formally, let $T_m = (\mathcal{V}_m,\mathcal{E}_m)$ be a linearly-parameterized, tree-structured pictorial structure for the $m$th mixture. Then, given image $\mathbf{I}$ and a face shape $\mathbf{s} = (\mathbf{x}_1,...,\mathbf{x}_N)^T$, the tree structured part model of view $m$ scores $\mathbf{s}$ as:
\begin{equation}\label{eq:tree_structured_part_model}
\begin{split}
  \mathcal{S}(\mathbf{I},\mathbf{s},m) &= \mathrm{App}_m(\mathbf{I},\mathbf{s}) + \mathrm{Shape}_m(\mathbf{s}) + \alpha^m \\
  \mathrm{App}_m(\mathbf{I},\mathbf{s}) &= \sum_{i \in \mathcal{V}_m} \mathbf{w}_i^m {\bm\cdot} \phi(\mathbf{I},\mathbf{x}_i) \\
  \mathrm{Shape}_m(\mathbf{s}) &= \sum_{ij \in \mathcal{E}_m} \! a_{ij}^{m}dx^2 \!+\! b_{ij}^{m}dx \!+\! c_{ij}^{m}dy^2 \!+\! d_{ij}^{m}dy,
\end{split}
\end{equation}
where $\mathrm{App}_m(\mathbf{I},\mathbf{s})$ sums the appearance evidence at each part in $\mathbf{s}$, $\mathrm{Shape}_m(\mathbf{s})$ scores the mixture-specific spatial arrangement of $\mathbf{s}$, and $\alpha^m$ is a scalar bias associated with view point mixture $m$. Since parts may look consistent across some changes in viewpoint, \cite{zhu2012face} allows different mixtures to share part templates to reduce the computational complexity.

To learn above mixtures of tree structured part models, the Chow-Liu algorithm \cite{chow1968approximating} is first used to find the maximum likelihood tree structure that best explains the face shape for a given mixture. Then, for each view, all the model parameters in Eq. \ref{eq:tree_structured_part_model} is discriminatively learned in a max-margin structured prediction framework. In the testing phase, the input image is scored by all tree structures $T_m = (\mathcal{V}_m,\mathcal{E}_m)$ respectively, and the globally optimal shape $\mathbf{s}$ is efficiently solved with dynamic programming algorithm \cite{felzenszwalb2005pictorial}.

Due to its simplicity and effectiveness, the tree structured part model \cite{zhu2012face} has been extensively investigated and improved for face alignment. U{\v{r}}i{\v{c}}{\'a}{\v{r}} \emph{et al.} \cite{uvrivcavr2016multi} argue that the learning algorithm of \cite{zhu2012face} is a variant of a two-class Support Vector Machines, which optimizes the detection rate of resulting face detector while the facial point locations serve only as latent variables not appearing in the loss function. In contrast, U{\v{r}}i{\v{c}}{\'a}{\v{r}} \emph{et al.} \cite{uvrivcavr2016multi} directly optimizes the average face alignment error with a novel objective function using the Structured Output SVMs algoirthm, which leads to a significant improvement in alignment accuracy. Yu \emph{et al.} \cite{yu2013pose} presented a two-stage cascaded deformable shape model for face alignment, where a group sparse learning method is proposed to automatically select the optimized anchor points to achieve robust initialization based on the part mixture model of \cite{zhu2012face}. Hsu \emph{et al.} \cite{hsu2015regressive} proposed to improve the run-time speed and localization accuracy of \cite{zhu2012face} with the Regressive Tree Structure Model (RTSM), where the tree structured model is applied on images with increasing resolution.

In general, the tree structured part model is effective at capturing global elastic deformation, while being easy to optimize unlike dense graph structure. Furthermore, it provide an unified framework to solve three tasks, namely face detection, face alignment and pose estimation, which is very appealing in automatic face analysis. However, its sluggish runtime impedes the potential for real-time facial point tracking; and perhaps due to the fact that the tree-based shape models allow for the non-face like structures to occur frequently, the performance of the tree structured part model \cite{zhu2012face} is reported to be slightly inferior to that of the CLMs \cite{saragih2011deformable,asthana2013robust}.

\vspace{0.25em}
\emph{A common limitation of above part-based discriminative methods (i.e., CLMs, constrained local regression, and DPMs), however, is that their performance is greatly constrained by the ambiguity of the local appearance models}. To break this bottleneck, many researchers have proposed to jointly estimate the whole face shape from the image, as described in the following sections.

\subsection{Ensemble regression-voting}
Apart from above local appearance model-based methods, another main stream of discriminative methods is to jointly estimate the whole face shape from image, during which the shape constraint is implicitly exploited. A simple way for this is to cast votes for the face shape from image patches via regression. Since voting from a single region is rather weak, a robust prediction is typically obtained by ensembling votes from different regions. We refer to these methods as \emph{ensemble regression-voting}. In general, the choice of the regression function, which can cast accurate votes for all facial points, is the key factor of the ensemble regression-voting approach.

Regression forests \cite{breiman2001random} are a natural choice to perform regression-voting due to their simplicity and low computational complexity. Cootes \emph{et al.} \cite{cootes2012robust} use random forest regression-voting to produce accurate response map for each facial point, which is then combined with the CLM fitting for robust prediction. Dantone \emph{et al.} \cite{dantone2012real} pointed out that conventional regression forests may lead to a bias to the mean face, because a regression forest is trained with image patches on the entire training set and averages the spatial distributions over all trees in the forest. Therefore, they extended the concept of regression forests to conditional regression forests. A conditional regression forest consists of multiple forests that are trained on a subset of the training data specified by global face properties (e.g., head pose used in \cite{dantone2012real}). During testing, the head pose is first estimated by a specialized regression forest, then trees of the various conditional forests are selected to estimate the facial points. Due to the high efficiency of random forests, \cite{dantone2012real} achieves close-to-human accuracy while processing images in real-time on the Labeled Faces in the wild (LFW) database \cite{huang2007labeled}. After that, Yang \emph{et al.} \cite{yang2013privileged} extended \cite{dantone2012real} by exploiting the information provided by global properties to improve the quality of decision trees, and later deployed a cascade of sieves to refine the voting map obtained from random regression forests \cite{yang2013sieving}. Apart from the regression forests \cite{dantone2012real,yang2012face,yang2013privileged,yang2013sieving}, Smith \emph{et al.} \cite{smith2014nonparametric} used each local feature surrounding the facial point to cast a weighted vote to predict facial point locations in a nonparametric manner, where the weight is pre-computed to take into account the feature's discriminative power.

In general, the ensemble regression-voting approach is more robust than previous local detector-based methods, and we conjecture that this robustness mainly stems from the combination of votes from different regions. However, current ensemble regression-voting approach, arguably, have not achieved a good balance between accuracy and efficiency for face alignment \emph{in-the-wild}. The random forests approach \cite{dantone2012real,yang2012face,yang2013privileged,yang2013sieving} is very efficient but can hardly cast precise votes for those unstable facial points (e.g., face contour), while on the other hand, the nonparametric feature voting approach based on facial part features \cite{smith2014nonparametric} is more accurate but suffers from very high computational burden. To pursue a face alignment algorithm that is both accurate and efficient, much research has focused on the cascaded regression approach as described in the next section.

\subsection{Cascaded regression} \label{sec:cascaded_regression}
Recently, cascaded regression has established itself as one of the most popular and state-of-the-art methods for face alignment, due to its high accuracy and speed \cite{cao2012face,sun2013deep,xiong2013supervised,ren2014face,zhu2015face}. The motivation behind this approach is that, since performing regression from image features to face shape in one step is extremely challenging, we can divide the regression process into stages, by learning a cascade of vectorial regressors.

Formally, given an image $\mathbf{I}$ and an initial shape $\mathbf{s}^0$, the face shape $\mathbf{s}$ is progressively refined by estimating a shape increment $\Delta \mathbf{s}$ stage-by-stage. In a generic form, a shape increment $\Delta \mathbf{s}$ at stage $t$ is regressed as:
\begin{equation}\label{eq:generic_cascaded_regression}
  \Delta \mathbf{s}^{t} = \mathcal{R}^{t}\big(\Phi^{t}(\mathbf{I},\mathbf{s}^{t-1})\big),
\end{equation}
where $\mathbf{s}^{t-1}$ is the shape estimated in the previous stage, $\Phi^{t}$ is the feature mapping function, and $\mathcal{R}^{t}$ is the stage regressor. Note that $\Phi^{t}(\mathbf{I},\mathbf{s}^{t-1})$ is referred to as \emph{shape-indexed} feature \cite{cao2012face,burgos2013robust} that depends on the current shape estimate, and can be either designed by hand \cite{xiong2013supervised,zhu2015face} or by learning \cite{cao2012face,ren2014face,kazemi2014one}. In the training phase, the stage regressors $(\mathcal{R}^1,...,\mathcal{R}^T)$ are sequentially learnt to reduce the alignment errors on training set, during which geometric constraints among points are \emph{implicitly} encoded.

\begin{figure*}[!htb]
\centering
\includegraphics[width=1\textwidth]{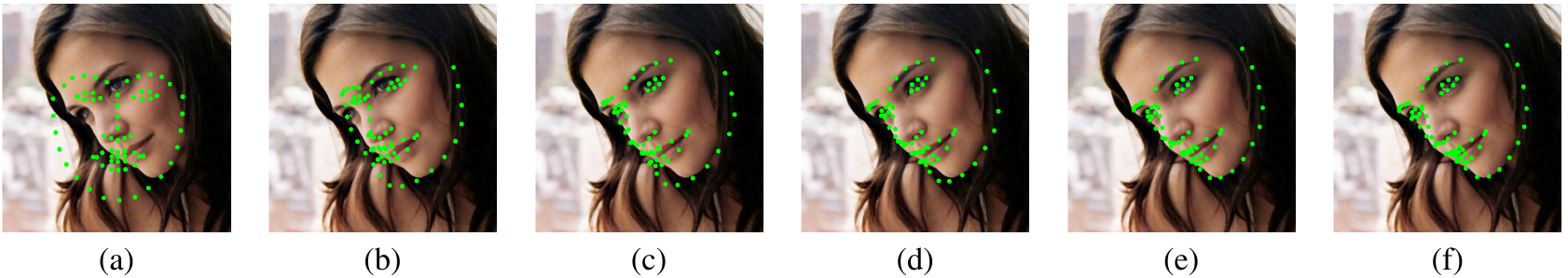}
\caption{Illustration of face alignment results in different stages of cascaded regression (Fig. 1 in \cite{lee2015face}). The shape estimate is initialized and iteratively updated through a cascade of regression trees: (a) initial shape estimate, (b)\text{-}(f) shape estimates at different stages.}
\label{fig:cascaded_regression}
\end{figure*}

%To learn the cascaded regressors $(\mathcal{R}^1,...,\mathcal{R}^T)$, a set of training samples $\{\mathbf{I}_i,\mathbf{s}_{i}^{*},\mathbf{s}_i^0\}_{i=1}^{D}$ are required, where $\mathbf{s}_i^{*}$ and $\mathbf{s}_i^0$ are the ground truth shape and the initial shape for image $\mathbf{I}_i$. In the training phase, the stage regressors $(\mathcal{R}^1,...,\mathcal{R}^T)$ are sequentially learnt to reduce the alignment errors on training set, during which geometric constraints among points are \emph{implicitly} encoded into the regressors. In particular, at stage $t$, the stage regressor $\mathcal{R}^t$ is formally learnt as follows,
%\begin{equation}\label{eq:stage_regressor_objective}
%  \mathcal{R}^t = \mathop{\mathrm{arg \, min}}_{\mathcal{R}} \sum_{i=1}^D ||\Delta \mathbf{s}_i^{t,*} - \mathcal{R}^{t}\big(\Phi^{t}(\mathbf{I},\mathbf{s}^{t-1})\big)||,
%\end{equation}
%where $\Delta \mathbf{s}_i^{t,*} = \mathbf{s}_i^{*} - \mathbf{s}_i^{t}$ is the shape residual. In \emph{testing}, given a test image $\mathbf{I}$ and an initial shape $\mathbf{s}^0$, the learned cascaded regressors $(\mathcal{R}^1,...,\mathcal{R}^T)$ are applied to compute shape increment and update the face shape in a cascade manner (see Fig. \ref{fig:cascaded_regression}).

Existing cascaded regression methods mainly differ in the specific form of the stage regressor $\mathcal{R}^t$ and the feature mapping function $\Phi^t$. Here, according to the type of the stage regressor $\mathcal{R}^t$, we roughly divide existing cascaded regression methods into two categories, i.e., \emph{two-level boosted regression}, and \emph{cascaded linear regression}.

\subsubsection{Two-level boosted regression} Cascaded regression is first introduced into face alignment by Cao \emph{et al.} \cite{cao2012face} in their seminal work called Explicit Shape Regression (ESR). They design a two-level boosted regression framework by again investigating boosted regression as the stage regressor $\mathcal{R}^t$. More specifically, they use a cascade of random ferns as $\mathcal{R}^t$ to regress the \emph{fixed} shape-indexed pixel difference feature at each stage, and adopt a correlation-based feature selection strategy to learn task-specific features. This combination makes ESR a break-through face alignment method in both accuracy and efficiency, and is widely adapted ever since.

Burgos-Artizzu \emph{et al.} \cite{burgos2013robust} also use the fern primitive regressor under the two-level boosted regression framework, but improve \cite{cao2012face} by explicitly incorporating the occlusion information into the regression target to better handle occlusions. Instead of random ferns used by \cite{cao2012face,burgos2013robust}, Kazemi \emph{et al.} \cite{kazemi2014one} present a general framework based on gradient boosting for learning an ensemble of regression trees, achieving super-realtime performance with high quality predictions and naturally handling missing or partially labelled data. Lee \emph{et al.} \cite{lee2015face} propose to use the Gaussian process regression tree (GPRT) to fit the primitive regressor under the two-level boosted regression framework, where GPRT is a Guassian process with a kernel defined by a set of trees.

\subsubsection{Cascaded linear regression} Although the two-level boosted regression framework has gained great success \cite{cao2012face,burgos2013robust,kazemi2014one,lee2015face}, generally speaking, any kind of stage regressor $\mathcal{R}^t$ with strong fitting capacity will be desirable. A notable example is the cascaded linear regression proposed by Xiong \emph{et al.} \cite{xiong2013supervised} using strong hand-craft SIFT \cite{lowe2004distinctive} feature.

The primary innovation of the cascaded linear regression method \cite{xiong2013supervised} is a Supervised gradient Descent Method (SDM) that gives a mathematically sound explanation of the cascaded linear regression by placing it in the context of Newton optimization for non-linear least squares problem. SDM shows that a Newton update for the non-linear least squares alignment error function can be expressed as a linear combination of the facial feature differences between the one extracted at current shape and the ground truth template, resulting in a linear update function $\mathcal{R}^{t}$ at each stage, i.e.,
\begin{equation}\label{eq:cascaded_linear_regression}
 \mathcal{R}^{t}:  \Delta \mathbf{s}^{t} \leftarrow \mathbf{W}^t \big(\Phi^{t}(\mathbf{I},\mathbf{s}^{t-1})\big) + \mathbf{b}^t,
\end{equation}
where $\Phi^{t}$ is the SIFT operator that extract SIFT feature at each facial point, and $\mathbf{W}^t$ is the \emph{averaged} descent direction on the training set.

Actually, SDM bears some similarities to AAMs trained in a discriminative manner with linear regression \cite{cootes2001active}, but differs from them in three aspects: (1) SDM is non-parametric in both shape and appearance; (2) SDM uses the part-based representation; (3) SDM learns different regressors $\mathcal{R}^t$ at different stages, while the original AAM \cite{cootes2001active} learns a constant regressor $\mathcal{R}$ for all stages.

Due to its concise formulation and state-of-the-art performance, SDM has been extensively investigated and extended. Xiong \emph{et al.} \cite{xiong2015global} pointed out that SDM is a local algorithm that is likely to average conflicting gradient directions, and proposed an extension of SDM called Global SDM (GSDM) that divides the search space into regions of similar gradient directions. Yan \emph{et al.} \cite{yan2013learn} proposed to generate multiple hypotheses, and then learn to rank or combine these hypotheses to get the final result. Asthana \emph{et al.} proposed an incremental formulation for the cascaded linear regression framework \cite{xiong2013supervised}, and presented multiple ways for incrementally updating a cascade of regression functions in an efficient manner. Zhu \emph{et al.} \cite{zhu2015face} designed a cascaded regression framework that begins with a coarse search over a shape space that contains diverse shapes, and employs the coarse solution to constrain subsequent finer search of shape, which improves the robustness of cascaded linear regression in coping with large pose variations.

\subsubsection{Discussion} Arguably, cascaded regression is playing a prominent role among the state-of-the-art methods for face alignment \emph{in-the-wild}. This is primarily because it has some distinct characteristics. (1) The training sample of cascaded regression is a triple defined by the face image, ground truth shape and the initial shape, which allows for convenient data augmentation by generating multiple initial shapes for one image. (2) It is capable of effectively leveraging large bodies of training data. (3) The shape constraints are encoded into regressors adaptively, which is more flexible than the parametric shape model that heuristically determines the model flexibility (e.g.,PCA dimension). (4) The cascaded regression framework is simple and generalizable, which allows different choices for the stage regressor $\mathcal{R}^t$ and convenient incorporation of feature learning techniques.

Although cascaded regression has achieved great success in face alignment, it is still not easy to perform regression from texture features to the whole shape update for some challenging faces with extreme expression or pose variation. This limitation can be partially confirmed by the fact that for some more flexible part localization task such as human pose estimation, the part detector-based methods still play a dominant role at present \cite{yang2013articulated,liu2015articulated}, rather than cascaded regression.

\subsection{Deep neural networks} \label{subsec:deep_neural_networks}
Deep neural networks, especially the deep convolutional network that can extract high-level image features, have been successfully utilized in many computer vision tasks, such as face verification \cite{taigman2014deepface,sun2014deep}, image classification \cite{krizhevsky2012imagenet,simonyan2014very,szegedy2015going}, and object detection \cite{girshick2014rich}. Naturally, they are also an effective choice to model the nonlinear relationship between the facial appearance and the face shape (or shape update).

However, applying deep network directly to face alignment is nontrivial due to the follwoing reasons: (1) While fine-tuning an existing CNN architecture (e.g., AlexNet \cite{krizhevsky2012imagenet}, GoogLeNet \cite{szegedy2015going}) to make it well adapted to the task at hand is very popular in computer vision \cite{girshick2014rich,zhang2014part}, such a strategy can hardly be applied for face alignment because the off-the-shelf large networks are typically trained for image classification while face alignment is a structural prediction problem. (2) Constructing a deep network-based system from scratch for face alignment should take into account the issue of over-fitting, and hence the network structures at each stage need to be carefully designed according to the task of this stage and the complexity involved.

\begin{figure*}[!htb]
\centering
\includegraphics[width=0.45\textwidth]{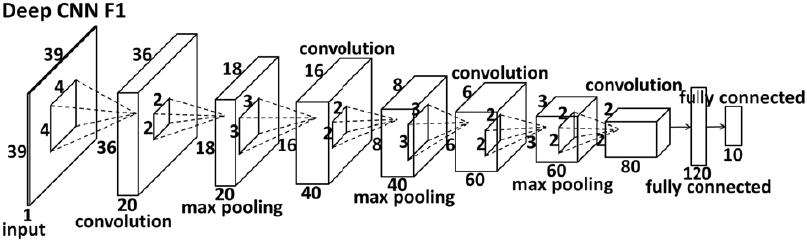}
\caption{One of the first-level convolutional neural network structures used in \cite{sun2013deep} to predict five major facial points. Sizes of input, convolution, and max pooling layers are illustrated by cuboids whose length, width, and height denote the number of maps, and the size of each map. Local receptive fields of neurons in different layers are illustrated by small squares in the cuboids.}
\label{fig:DCNN}
\end{figure*}

Focusing on above issues, Sun \emph{et al.} \cite{sun2013deep} were pioneers in this area with their work called Deep Convolutional Network Cascade. They handled the face alignment task with three-level carefully designed convolutional networks, and fuse the outputs of multiple networks at each level for robust prediction (Fig. \ref{fig:DCNN} illustrates one of the first-level CNN structures). The first level network takes the whole face image as input to predict the initial estimates of the holistic face shape, during which the shape constraints are implicitly encoded. Then, the following two level networks refine the position of each point to achieve higher accuracy. Several network structures critical for face alignment are investigated in \cite{sun2013deep}, providing some important principles on the choice of convolutional network structures. For example, convolutional networks at the first level should be deeper than the following networks, since predicting facial points from large input regions is a high-level task.

Ever since the work of \cite{zhang2014facial}, deep CNNs have been widely exploited for face alignment. Similar to \cite{zhang2014facial}, Zhou \emph{et al.} \cite{zhou2013extensive} designed a four-level convolutional network cascade to tackle the face alignment problem in a coarse-to-fine manner, where each network level is trained to locally refine a subset of facial points generated by previous network levels. Zhang \emph{et al.} \cite{zhang2014facial} extended the work of \cite{sun2013deep} by jointly learning auxiliary attributes along with face alignment. Their work confirms that some heterogeneous but subtly correlated tasks, e.g., head pose estimation and facial attribute inference can aid the face alignment task through multi-task learning. Lai \emph{et al.} \cite{lai2015deep} proposed an end-to-end CNN architecture to learn highly discriminative shape-indexed features, by encoding the image into high-level feature maps in the same size of the image, and then extracting deep features from these high level descriptors through a novel ``Shape-Indexed Pooling" method.
Despite of the great popularity and success, as mentioned before, we should take into account the tradeoff between the model complexity and training data size, since some deep models have been reported to be pre-trained with enormous quantity of external data sources \cite{sun2013deep,zhang2014facial}.

\subsection{Summary and discussion}
We have reviewed discriminative methods for face alignment in six groups, i.e., \emph{CLMs}, \emph{constrained local regression}, \emph{DPMs}, \emph{ensemble regression-voting}, \emph{cascaded regression} and \emph{deep neural networks}. Among them, CLMs, constrained local regression and DPMs follow the ``divide and conquer" principle to simplify the face alignment task by constructing individual local appearance model for each facial point. However, due to their small patch support and large appearance variation in training, these local appearance models are typically plagued by the problem of ambiguity. Furthermore, since further inference (or global shape optimization) is based on the detection responses of these local appearance models, the problem of ambiguity may create the most serious performance bottleneck for face alignment \emph{in-the-wild}.

To break this bottleneck, another main stream in face alignment is to jointly estimate the whole face shape from image, implicitly exploiting the spatial constraints among facial points. In this line, we have first reviewed the \emph{ensemble regression-voting} and \emph{cascaded regression} methods, which learn a vectorial regression function to infer the whole face shape in an ensemble or cascaded manner. In particular, cascaded regression has emerged as one of the most popular and state-of-the-art methods, due to its speed, accuracy and robustness. Then, we briefly reviewed the deep learning-based approach for face alignment, which have the advantage of learning highly discriminative task-specific features, but should take into account the issue of over-fitting.

It is worth noting that some methods involve techniques motivated by different principles, which clearly overlap our category boundaries. For example, we classify the regression voting-based shape model matching method \cite{cootes2012robust} as CLM, since they fit a parametric shape model to a new image based on the response map for each facial point. However, since the response maps in \cite{cootes2012robust} are generated by random forest regression-voting, it can also be considered as an ensemble regression-voting method. Furthermore, some deep learning-based methods can also be classified as cascaded regression due to their cascaded structure \cite{zhang2014coarse,lai2015deep}.

\section{Towards the development of a robust face alignment system}\label{sec_development}
Face alignment \emph{in-the-wild} is very challenging due to many kinds of undesirable appearance variations, and hence it is often the case that no single modality is enough. In this section, we will focus on the practical aspects of constructing a robust face alignment system, which is mostly ignored in previous studies. Specifically, we first present a global system architecture for face alignment, and then have a close look at possible strategies to improve the robustness of face alignment under this architecture.

\subsection{The global system architecture for face alignment}
\begin{figure*}[!htb]
\centering
\includegraphics[width=0.95\textwidth]{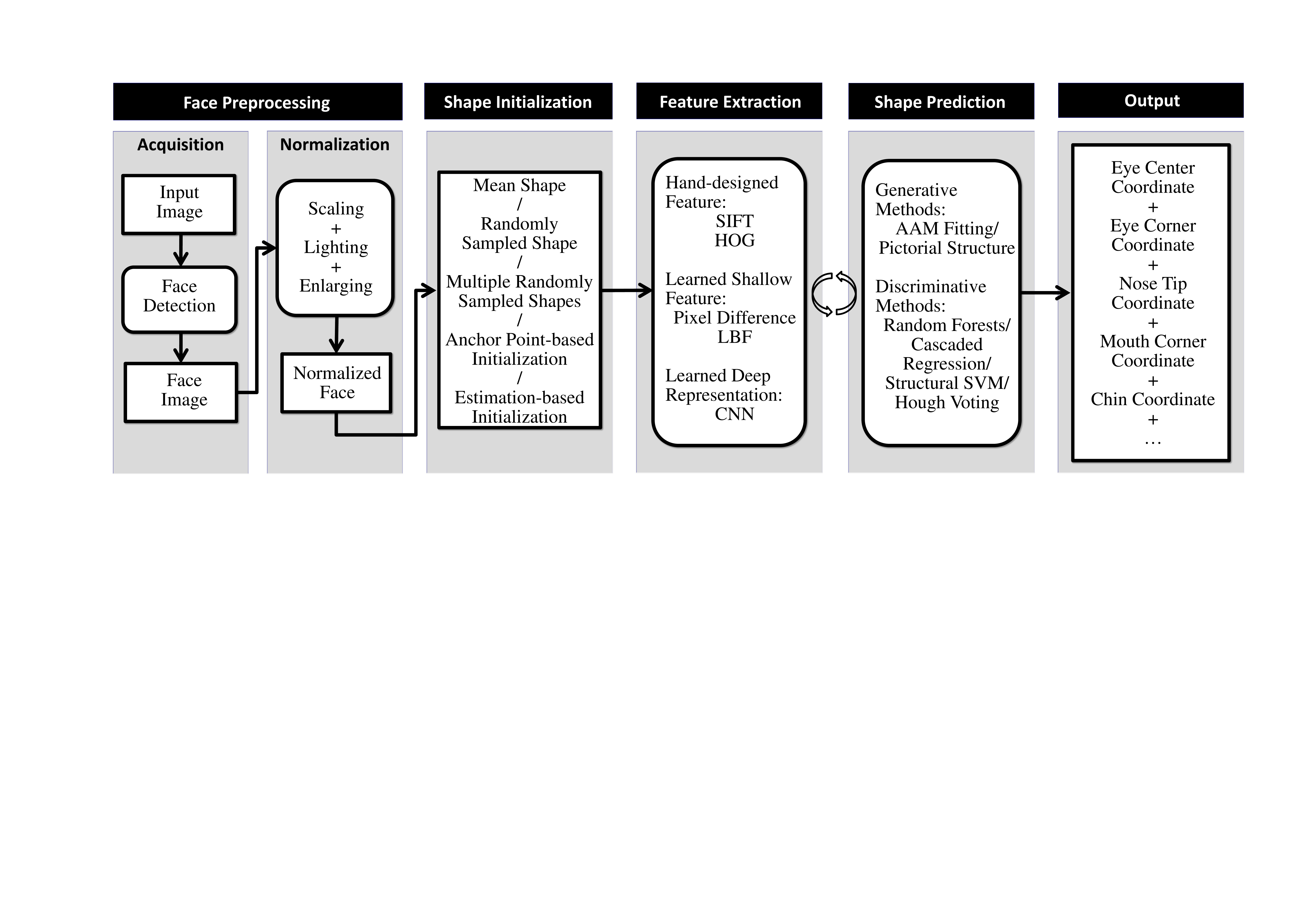}
\caption{A global system architecture for face alignment.}
\label{fig_deve_architecture}
\end{figure*}
Inspired by \cite{song2013literature,fasel2003automatic}, we give a global system architecture for face alignment, where a complicated system is divided into several substages. As shown in Fig. \ref{fig_deve_architecture}, the architecture can be roughly divided into three parts: face preprocessing, shape initialization, and the iterative process of feature extraction and shape prediction. We note that this architecture is only to illustrate a general pipeline for face alignment, while in practical not all components are mandatory. For example, the consensus of exemplar method \cite{belhumeur2011localizing} do not involve the shape initialization step.

While the feature extraction and shape prediction process have drawn a great deal of attention in literature, the face preprocessing and shape initialization steps are often ignored. Meanwhile, problems such as training data augmentation, and the accuracy and efficiency tradeoff are also essential for any practical face alignment system. In the following, we will have a closer look at these issues.

\subsection{Training data augmentation}
Due to the difficulty and cost of manual annotation, the number of training samples we \emph{actually} have is often much smaller than that we \emph{supposedly} have. In such a case, artificial data augmentation, which is usually done by label-preserving transforms, is the easiest and most common method to reduce over-fitting.

In general, there are four distinct forms of data augmentation to enlarge the training set: (1) generating image rotations from a small interval (e.g., [-15 degrees, +15 degrees] used in \cite{belhumeur2013localizing}); (2) synthesizing images by left-right flip to double the training set; (3) disturbing the bounding boxes by randomly scaling and translating the bounding box for each image, which also increases the robustness of face alignment algorithms to the bounding boxes; (4) sampling multiple initialization for each training image, which is typically used by cascaded regression methods.

\subsection{Face preprocessing} \label{subsec_development_face_preprocessing}
For the task of face alignment, it is useful to remove the scaling variations of the detected faces, and enlarge the face region to ensure that all predefined facial points are enclosed.

\subsubsection{Handling scaling variations} Typically, for a face analysis system, the training and test faces are required to be roughly the same scale, by rescaling the bounding box produced by the face detector. We note that to help preserve more detailed texture information, the size of the normalized bounding box for high-resolution face databases is typically chosen to be larger than that for low-resolution face databases. For example, Belhumeur \emph{et al.} \cite{belhumeur2013localizing} rescale the high-resolution images from the LFPW database so that the faces have an inter-ocular distance of roughly 55 pixels, while Dantone \emph{et al.} \cite{dantone2012real} choose to rescale the bounding box of the low-resolution faces from the LFW database \cite{huang2007labeled} to 100$\times$100, which is slightly smaller than the size chosen by Belhumeur \emph{et al.} \cite{belhumeur2013localizing}.

\subsubsection{Enlarging face areas} The output of a face detector is a rough face region that might miss some facial points (e.g., the chin). This has little impact on cascaded regression, for which the bounding box only serves to rescale the face and compute the initial shape. However, for those methods based on exhaustive search or feature voting, it is necessary to enlarge the face bounding box to enclose all the facial points, or define the sampling region of image patches to cast votes. For this, Dantone \emph{et al.} \cite{dantone2012real} suggest to enlarge the face bounding box by 30\%, and we believe that this strategy may satisfy the requirements of all face alignment algorithms.

\subsection{Shape initialization}
Most face alignment methods start from a rough initialization, and then refine the shape iteratively until convergence. The initialization step typically has great influence on the final result, and an initial shape far from the ground truth might lead to very bad alignment results.

The most common choice is to use the \emph{mean shape} for initialization \cite{xiong2013supervised,kazemi2014one,ren2014face}. However, sometimes, the mean shape is likely to be far from the target shape, and leads to bad result. As an alternative, Cao \emph{et al.} \cite{cao2012face} propose to run the algorithm several times using different initialisations \emph{randomly} sampled from the training shapes, and take the median result as the final estimation to improve robustness. Burgos-Artizzu \emph{et al.} \cite{burgos2013robust} proposed a smart restart method to further improve the multiple initialization strategy in \cite{cao2012face} by checking the variance between the predictions using different initializations.

Recently, some authors proposed to estimate an initial shape that is tailored to the input face. Zhang \emph{et al.} \cite{zhang2014facial} showed that the five major facial points localized by their deep model can serve as anchor points to apply similarity transform to randomly sampled training shapes. Through this, very accurate initial shapes can be generated for other algorithms (e.g., \cite{burgos2013robust}) and lead to promising performance improvement. Zhang \emph{et al.} \cite{zhang2014coarse} and Sun \emph{et al.} \cite{sun2013deep} proposed to directly estimate a rough initial shape from the global image, which in general produces good initial shape that aids following alignment.

\subsection{Accuracy and efficiency tradeoffs}
Face alignment in real time is crucial to many practical applications. The efficiency mainly depends on the feature extraction and shape prediction steps. In general, strong hand-designed feature (e.g., SIFT \cite{lowe2004distinctive}) captures detailed texture information that may aid detection, but have higher computational cost compared to simpler features (e.g., BRIEF \cite{calonder2010brief}). Zhu \emph{et al.} \cite{zhu2015face} identified this phenomenon under the cascaded regression framework, and proposed to exploit different types of features at different stages to achieve a good trade-off between accuracy and efficiency, i.e., employ less accurate but computationally efficient BRIEF feature at the early stages, and use more accurate but relatively slow SIFT feature at later stages. Besides this hybrid strategy, a better choice is to learn highly efficient and discriminative features \cite{cao2012face,ren2014face,kazemi2014one}. In particular, Ren \emph{et al.} \cite{ren2014face} propose to learn a set of highly discriminative local binary features for each facial point independently. Because extracting and regressing local binary features is computationally very cheap, \cite{ren2014face} achieves over 3,000 FPS while obtaining accurate alignment result.

In term of shape prediction, the regression-based methods in general are very efficient, while the exhaustive search based methods typically suffer from high computational cost \cite{belhumeur2011localizing,zhou2013exemplar}. Dibeklio{\u{g}}lu \emph{et al.} \cite{dibekliouglu2012statistical} propose to mitigate this issue through a coarse-to-fine search strategy. In \cite{dibekliouglu2012statistical}, a three-level image pyramid from the cropped high-resolution face images is designed to reduce the search region, where the coarse-level images have lower resolution but much smaller size.

\section{System Evaluation}\label{sec_evaluation}
In this section, we first review the major wild face databases and evaluation metric in the literature, then summarize and discuss some of reported performance of current state-of-the-art, on the several popular wild face databases using the same evaluation metric for reference.

\subsection{Databases and metric}

\subsubsection{Databases}
\begin{table}[!htb]
  \centering
  \scriptsize
  \caption{A list of sources of wild databases for face alignment.}
  \label{tab_evaluation_database}
  \begin{threeparttable}
  \begin{tabular}{p{0.098\textwidth}p{0.03\textwidth}<{\centering}p{0.065\textwidth}<{\centering}p{0.08\textwidth}<{\centering}p{0.045\textwidth}<{\centering}p{0.045\textwidth}<{\centering}p{0.485\textwidth}}
  \Xhline{1pt}
  \bf{Databases} &\bf{Year} &\bf{\#Images} &\bf{\#Training} &\bf{\#Test} & \bf{\#Point} & \bf{Links}\\
  \Xhline{0.5pt}
  LFW \cite{huang2007labeled} & 2007 & 13,233 & 1,100 & 300 & 10 & \url{http://www.dantone.me/datasets/facial-features-lfw/}\\
  LFPW \cite{belhumeur2011localizing} & 2011 & 1,432\tnote{a} & - & - & 35\tnote{b} & \url{http://homes.cs.washington.edu/~neeraj/databases/lfpw/}\\
  AFLW \cite{kostinger2011annotated} & 2011 & 25,993 & - & - & 21 & \url{http://lrs.icg.tugraz.at/research/aflw}\\
  AFW \cite{zhu2012face} &2012 & 205 & - & - & 6 & \url{http://www.ics.uci.edu/~xzhu/face/}\\
  HELEN \cite{le2012interactive} &2012 & 2,330 & 2,000 & 300 & 194 & \url{http://www.ifp.illinois.edu/~vuongle2/helen/}\\
  300-W \cite{sagonas2013semi} & 2013 &3,837 & 3,148 & 689 & 68 &\url{http://ibug.doc.ic.ac.uk/resources/300-W/}\\
  COFW \cite{burgos2013robust} & 2013 &1,007 & - & - & 29 &\url{http://www.vision.caltech.edu/xpburgos/ICCV13/}\\
  MTFL \cite{zhang2014facial} & 2014 & 12,995 & - & - & 5 &\url{http://mmlab.ie.cuhk.edu.hk/projects/TCDCN.html}\\
  MAFL \cite{zhang2016learning} & 2016 &20,000 & - & - & 5 & \url{http://mmlab.ie.cuhk.edu.hk/projects/TCDCN.html}\\
  \Xhline{1pt}
  \end{tabular}
  \begin{tablenotes}
    \tiny
    \item[a] LFPW is shared by web URLs, but some URLs are no longer valid.
    \item[b] Each face image in LFPW is annotated with 35 points, but only 29 points defined in \cite{belhumeur2011localizing} are used for the face alignment.
  \end{tablenotes}
  \end{threeparttable}
\end{table}

\begin{figure}[!htb]
\centering
\includegraphics[width=0.6\textwidth]{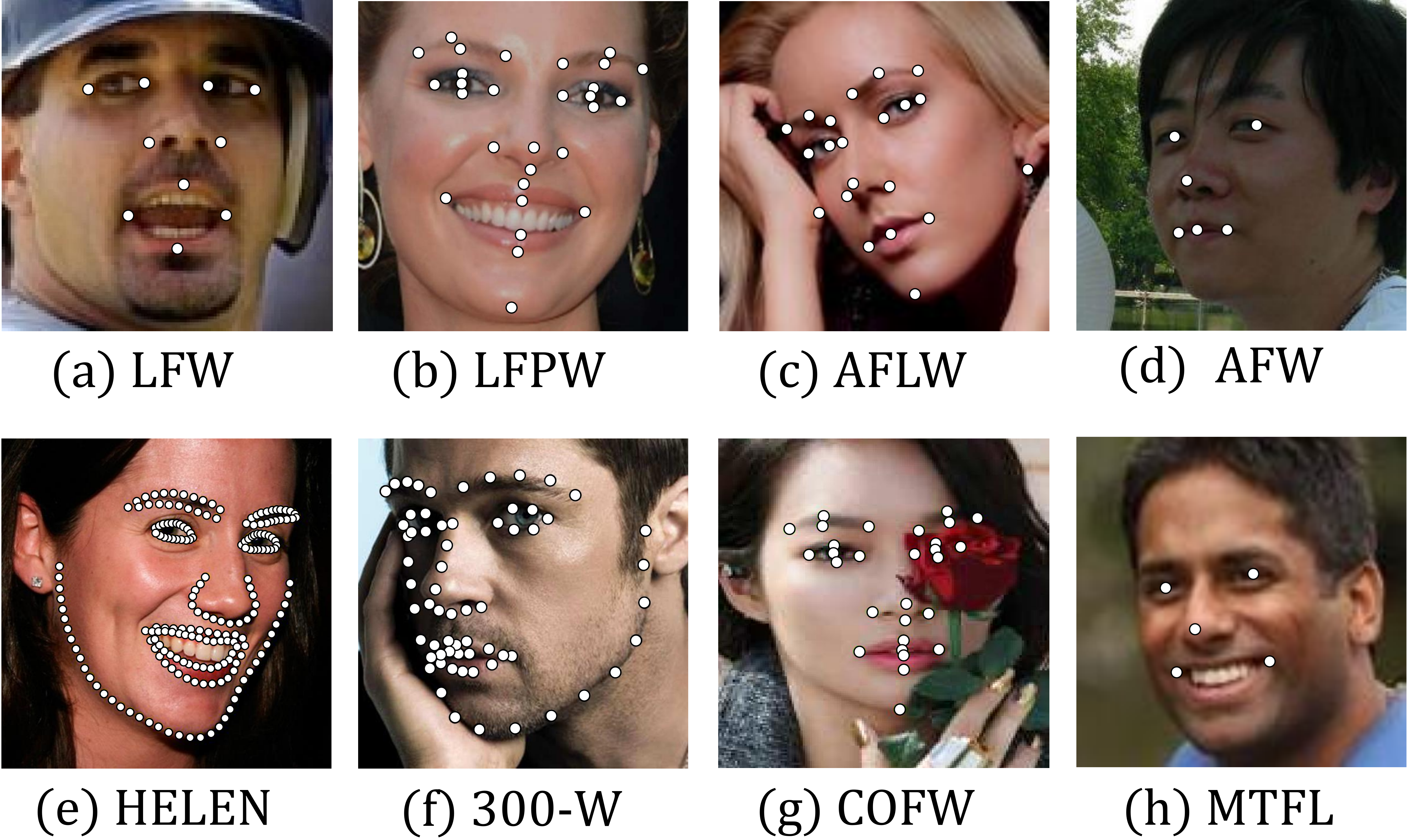}
\caption{Illustration of the example face images from eight wide face databases with original annotation.}
\label{fig_evaluation_annotaion}
\end{figure}

There have been many face databases developed for face alignment, with the ground truth facial points labelled manually by employing workers or through the tools such as Amazon mechanical turk (MTurk). Among them, some databases are collected under controlled laboratory conditions with normal lighting, neutral expression and high image quality, including the Extended M2VTS database (XM2VTS) \cite{messer1999xm2vtsdb}, BioID face database \cite{jesorsky2001robust}, PUT \cite{kasinski2008put}, Multi-Pie \cite{gross2010multi}, etc.

However, the goal of this paper is to investigate the problem of face alignment \emph{in-the-wild}, so we are more concerned with the \emph{uncontrolled} databases that exhibit large facial variations due to pose, expressions, lighting, occlusion and image quality. These uncontrolled databases are typically collected from social network such as google.com, flickr.com, facebook.com, which are more realistic and challenging for face alignment. In Tab. \ref{tab_evaluation_database}, we list the basic information of 9 wild face databases, including LFW \cite{huang2007labeled}, LFPW \cite{belhumeur2011localizing}, AFLW \cite{kostinger2011annotated}, AFW \cite{zhu2012face}, HELEN \cite{le2012interactive}, 300-W \cite{sagonas2013semi}, COFW \cite{burgos2013robust}, MTFL \cite{zhang2014facial}, and MAFL \cite{zhang2016learning}, and also provide links to download them. The example face images from these databases with original annotation are illustrated in Fig. \ref{fig_evaluation_annotaion}. It is worth noting that the LFPW, AFW and HELEN databases are re-annotated by Sagonas \emph{et al.} \cite{sagonas2013semi} with 68 points.

\subsubsection{Evaluation metric} There have been several evaluation metrics for the alignment accuracy in the literature. For example, many authors reported the inter-pupil distance normalized facial point error averaged over all facial points and images for each database \cite{burgos2013robust,ren2014face,kazemi2014one,zhu2015face,lee2015face}. Specifically, the inter-ocular distance normalized error for facial point $i$ is defined as:
\begin{equation}\label{eq_evaluation_metric}
  e_i = \frac{||\mathbf{x}_i-\mathbf{x}_i^*||_2}{d_{IOD}},
\end{equation}
where $\mathbf{x}_i$ is the automatically localized facial point location, $\mathbf{x}_i^*$ is the manually annotated location, and $d_{IOD}$ is the inter-ocular distance. The normalization term $d_{IOD}$ in this formulation can eliminate unreasonable measurement variations caused by variations of face scales.

The cumulative errors distribution (CED) curve is also often chosen to illustrate the comparative performance, showing the proportion of the test images or facial points with the increase of the normalized error \cite{saragih2011deformable,belhumeur2011localizing,tzimiropoulos2014gauss,tzimiropoulos2015project,zhu2015face}. Some other evaluation metric can also been found in literature, such as the facial point error normalized by face size \cite{yu2013pose}, the percentage of the test images or facial points less than given relative error level \cite{dibekliouglu2012statistical,yu2013pose}, and the percentage of accuracy improvement over other algorithm \cite{cao2012face}.

Besides the accuracy, the efficiency is another important performance indicator of face alignment algorithms, which is typically measured by frames per second (FPS).

\subsection{Evaluation and discussion}
\begin{table*}[!htb]
  \centering
  \tiny
  \caption{A list of published software of face alignment.}
  \label{tab_evaluation_software}
  \begin{tabular}{p{0.4\textwidth}p{0.05\textwidth}<{\centering}p{0.05\textwidth}<{\centering}p{0.4\textwidth}}
  \Xhline{1pt}
  \bf{Methods} & \bf{Year} & \bf{$\#$Points} & \bf{Links}\\
  \Xhline{0.5pt}
  Boosted Regression with Markov Networks (BoRMaN) \cite{valstar2010facial} & 2010 & 22 & \url{http://ibug.doc.ic.ac.uk/resources/facial-point-detector-2010/}\\
  Constrained Local Model (CLM) \cite{saragih2011deformable} & 2011 & 66 & \url{https://github.com/kylemcdonald/FaceTracker}\\
  Tree Structured Part Model (TSPM) \cite{zhu2012face} & 2012 & 68 & \url{http://www.ics.uci.edu/~xzhu/face/}\\
  Conditional Random Forests (CRF) \cite{dantone2012real} & 2012 & 10 & \url{http://www.dantone.me/projects-2/facial-feature-detection/}\\
  Structured Output SVM \cite{uvrivcavr2012detector} & 2012 & 7 & \url{http://cmp.felk.cvut.cz/~uricamic/flandmark/}\\
  Cascaded CNN \cite{sun2013deep} & 2013 & 5 & \url{http://mmlab.ie.cuhk.edu.hk/archive/CNN_FacePoint.htm}\\
  Discriminative Response Map Fitting (DRMF) \cite{asthana2013robust} & 2013 & 66 & \url{https://sites.google.com/site/akshayasthana/clm-wild-code?}\\
  Supervised Descent Method (SDM) \cite{xiong2013supervised} & 2013 & 49 & \url{www.humansensing.cs.cmu.edu/intraface}\\
  Robust Cascaded Pose Regression (RCPR) \cite{burgos2013robust} & 2013 & 29 & \url{http://www.vision.caltech.edu/xpburgos/ICCV13/}\\
  Optimized Part Mixtures (OPM) \cite{yu2013pose} & 2013 & 68 & \url{http://www.research.rutgers.edu/~xiangyu/face_align/face_align_iccv_1.1.zip}\\
  Continuous Conditional Neural Fields (CCNF) \cite{baltruvsaitis2014continuous} & 2014 & 68 & \url{https://github.com/TadasBaltrusaitis/CCNF}\\
  Coarse-to-fine Shape Searching (CFSS) \cite{zhu2015face} & 2015 & 68 & \url{mmlab.ie.cuhk.edu.hk/projects/CFSS.html}\\
  Project-Out Cascaded Regression (PO-CR) \cite{tzimiropoulos2015project} & 2015 & 68 & \url{http://www.cs.nott.ac.uk/~yzt/}\\
  Active Pictorial Structures (APS) \cite{antonakos2015active} & 2015 & 68 & \url{https://github.com/menpo/menpo}\\
  Tasks-Constrained Deep Convolutional Network (TCDCN) \cite{zhang2016learning} & 2016 & 68 & \url{http://mmlab.ie.cuhk.edu.hk/projects/TCDCN.html}\\
  \Xhline{1pt}
  \end{tabular}
\end{table*}

\begin{figure*}[!htb]
\centering
\includegraphics[width=0.98\textwidth]{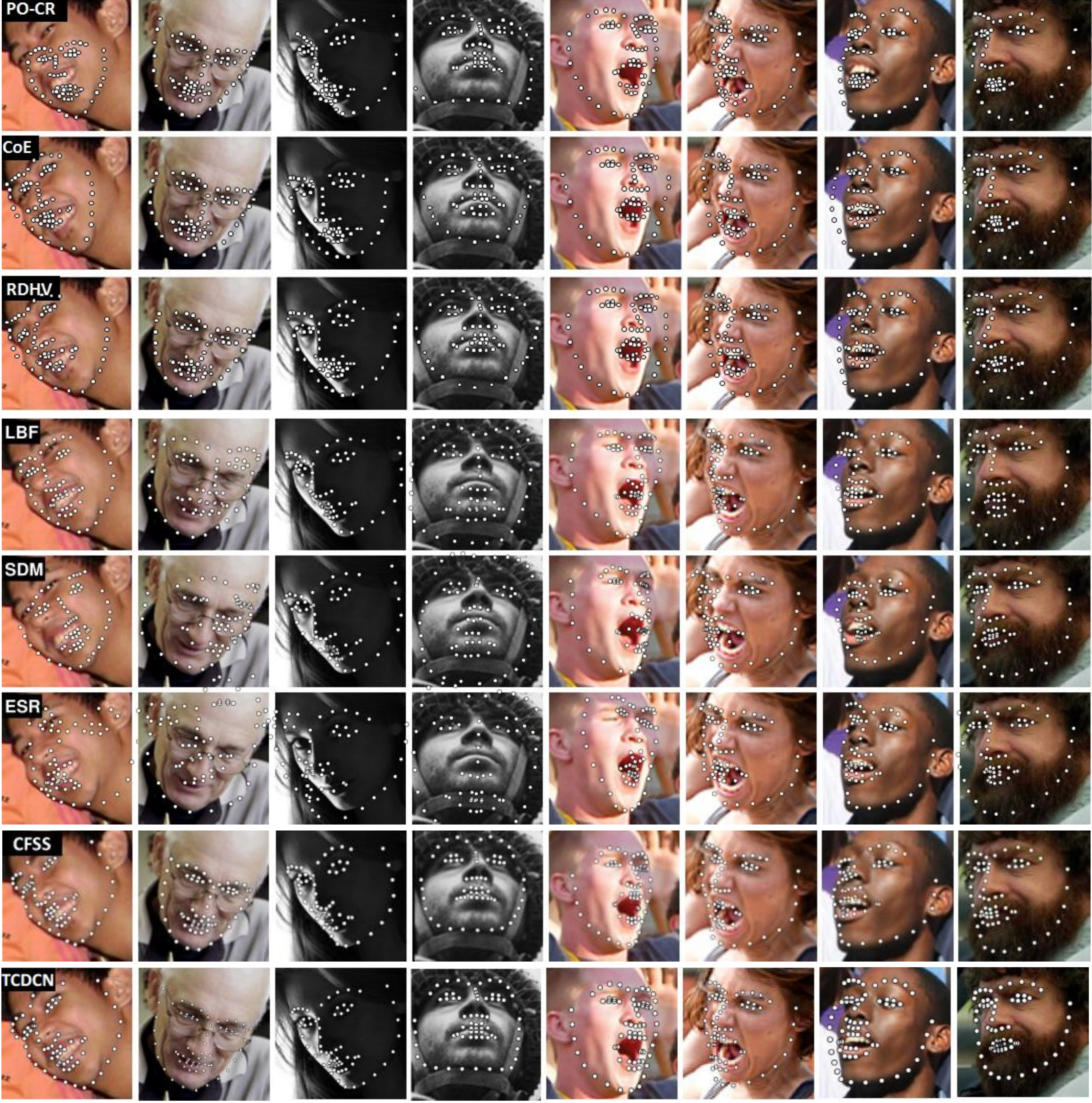}
\caption{Example results on IBUG database \cite{sagonas2013300} by eight state-of-the-art methods. These images are
extremely difficult due to the mixing of large head poses, extreme lighting, and partial occlusions. From top to bottom, results are produced by the Project-Out Cascaded Regression (PO-CR) method \cite{tzimiropoulos2015project}, Consensus of Exemplar (CoE) method \cite{belhumeur2013localizing}, Robust Discriminative Hough Voting (RDHV) method \cite{jin2016face}, Local Binary Feature (LBF) method \cite{ren2014face}, Supervised Descent Method (SDM) \cite{xiong2013supervised}, Explicit Shape Regression (ESR) method \cite{cao2012face}, Coarse-to-Fine Shape Searching (CFSS) method \cite{zhu2015face}, Tasks-Constrained Deep Convolutional Network (TCDCN) method \cite{zhang2016learning}. Among these methods, we implement the Consensus of Exemplar (CoE) \cite{belhumeur2013localizing} and Robust Discriminative Hough Voting (RDHV) \cite{jin2016face} methods and test them on these images, while other results are obtained from the published papers.}
\label{fig:evaluation_alignment_examples}
\end{figure*}

We choose four common wild databases, i.e., LFW, LFPW, HELEN, 300-W and IBUG (challenging subset of 300-W) databases, to show comparative performance statistics of the state of the art. Table \ref{tab_evaluation_software} lists some softwares published online, and Table \ref{tab_evaluation} summarizes the reported performance on above databases. Fig. \ref{fig:evaluation_alignment_examples} shows some challenging images from IBUG aligned by eight state-of-the-art methods respectively.

For performance evaluation, we are mainly concerned with two key performance indicators, i.e., accuracy and efficiency. The former is measured by the normalized facial point error (cf. Eq. \ref{eq_evaluation_metric}) averaged over all facial points and images for each database, while the later is measured by frames per second (FPS).

\begin{figure}[!htb]
\centering
\subfigure[] {\label{fig:error_tolerance_10}
\includegraphics[width=0.175\textwidth]{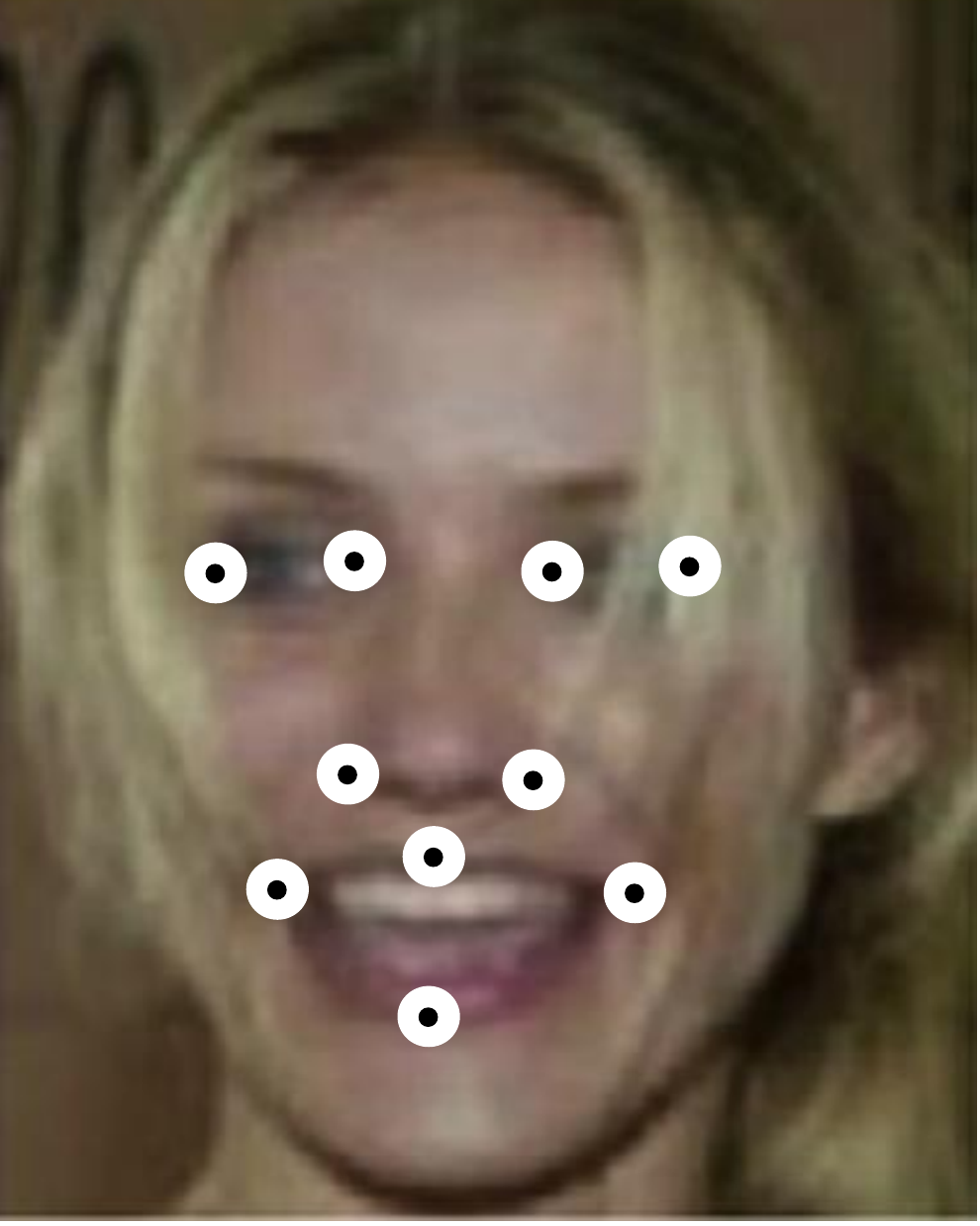}
}
\subfigure[] {\label{fig:error_tolerance_5}
\includegraphics[width=0.175\textwidth]{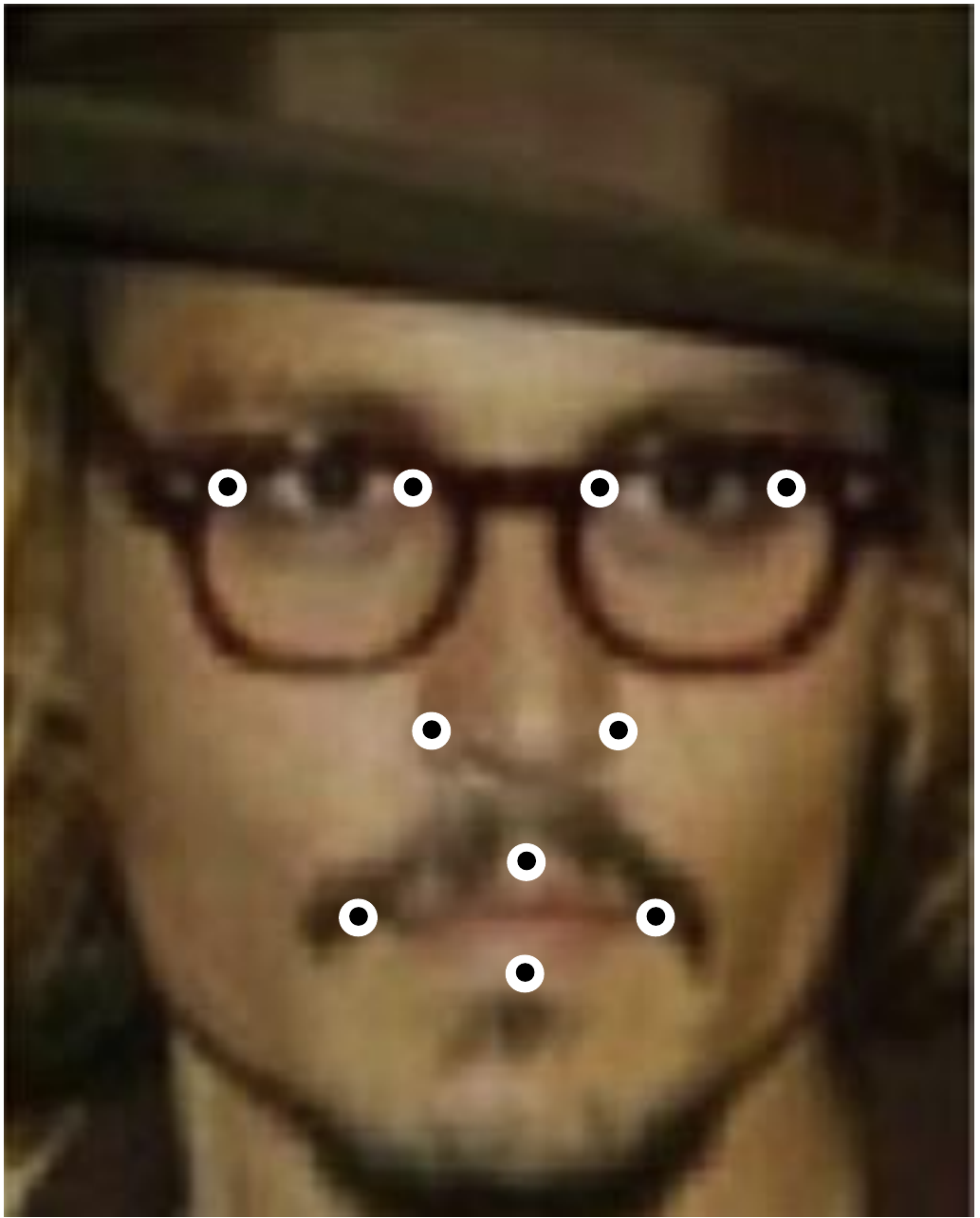}
}
\caption{Two images of the LFW database annotated with 10 facial feature points. The white circles show the disturbance range from the ground truth (black points), 10\% of the inter-ocular distance in (a) while 5\% in (b), which aims to give a intuitive feeling of the localization error listed in Table \ref{tab_evaluation}. }
\label{fig:error_tolerance}
\end{figure}

%\begin{figure*}[!htb]
%\centering
%\includegraphics[width=1\textwidth]{images/evaluation/LBF_good_cases.pdf}
%\caption{Example results on the IBUG database \cite{sagonas2013300} by Local Binary Feature (LBF) method \cite{ren2014face} (Fig. 7 in \cite{ren2014face}).}
%\label{fig:evaluation_good_cases}
%\end{figure*}

\begin{table*}[!htb]
\tiny
\centering
\caption{Lists of face alignment performance evaluated on various wild face databases.}
\label{tab_evaluation}
\begin{threeparttable}
\begin{tabular}{p{0.06\textwidth}p{0.15\textwidth}p{0.06\textwidth}p{0.06\textwidth}p{0.35\textwidth} p{0.08\textwidth}p{0.1\textwidth}}
\Xhline{1pt}
\bf{Databases}& \bf{Challenges} & \bf{$\#$Test} & \bf{$\#$Points} &  \bf{Methods} &  \bf{Error (\%)} & \bf{FPS}\\
\Xhline{0.5pt}
\multirow{5}{0.06\textwidth}{LFW \cite{huang2007labeled}} & \multirow{5}{0.15\textwidth}{Low resolution, large variations in illuminations, expressions and poses}  & \renewcommand{\multirowsetup}{\centering} \multirow{5}{0.06\textwidth}{13,233\tnote{a}} & \renewcommand{\multirowsetup}{\centering}  \multirow{4}{0.06\textwidth}{10} & Conditional Random Forests (CRF) \cite{dantone2012real} & 7.00 & 10 (c++)\\
& & & & Explicit Shape Regression (ESR) \cite{cao2012face} & 5.90 & 11 (Matlab)\\
& & & & Robust Cascaded Pose Regression (RCPR) \cite{burgos2013robust}  & 5.30 & 15 (Matlab)\\
& & & & & & \\
& & & \renewcommand{\multirowsetup}{\centering} \multirow{1}{0.06\textwidth}{55\tnote{b}}  & Consensus of Exemplar (CoE) \cite{belhumeur2013localizing} & 5.18 & - \\

\Xhline{0.5pt}
\multirow{16}{0.06\textwidth}{LFPW \cite{belhumeur2011localizing}} &\multirow{16}{0.15\textwidth}{Large variations in illuminations, expressions, poses and occlusion}  & \renewcommand{\multirowsetup}{\centering} \multirow{16}{0.06\textwidth}{224$\sim$300\tnote{c}} & \renewcommand{\multirowsetup}{\centering} \multirow{7}{0.06\textwidth}{21} &  Consensus of Exemplar (CoE)  \cite{belhumeur2011localizing}  & 3.99 & $\approx$ 1 (C++)\\
& & & & Explicit Shape Regression (ESR) \cite{cao2012face} & 3.47 & 220 (C++)\\
& & & & Robust Cascaded Pose Regression (RCPR) \cite{burgos2013robust} & 3.50 & 12 (Matlab)\\
& & & & Supervised Descent Method (SDM) \cite{xiong2013supervised} & 3.49 & 160 (C++)\\
& & & & Exemplar-based Graph Matching (EGM) \cite{zhou2013exemplar} & 3.98 & $<$ 1\\
& & & & Local Binary Feature (LBF) \cite{ren2014face} & 3.35 & 460 (C++)\\
& & & & Fast Local Binary Feature (LBF fast) \cite{ren2014face} & 3.35 & 4600 (C++)\\
& & & & & & \\
& & & \renewcommand{\multirowsetup}{\centering} \multirow{8}{0.06\textwidth}{68\tnote{d}}  & Tree Structured Part Model (TSPM)  \cite{zhu2012face} & 8.29 & 0.04 (Matlab) \\
& & & & Discriminative Response Map Fitting (DRMF) \cite{asthana2013robust} & 6.57 & 1 (Matlab)\\
& & & & Robust Cascaded Pose Regression (RCPR) \cite{burgos2013robust} & 6.56 & 12 (Matlab)\\
& & & & Supervised Descent Method (SDM) \cite{xiong2013supervised} & 5.67 & 70 (C++)\\
& & & & Gauss-Newton Deformable Part Model (GN-DPM) \cite{tzimiropoulos2014gauss} & 5.92 & 70\\
& & & & Coarse-to-fine Auto-encoder Networks (CFAN) \cite{zhang2014coarse} & 5.44 & 20 \\
& & & & Coarse-to-fine Shape Searching (CFSS) \cite{zhu2015face} & 4.87 & -\\
& & & & CFSS Practical \cite{zhu2015face} & 4.90 & -\\
& & & & Deep Cascaded Regression (DCR) \cite{lai2015deep} & 4.57 & -\\
\Xhline{0.5pt}
\multirow{20}{0.06\textwidth}{HELEN \cite{le2012interactive}} & \multirow{20}{0.15\textwidth}{Computation burden due to the dense annotation, large variations in expressions, poses and occlusion}  & \renewcommand{\multirowsetup}{\centering} \multirow{20}{0.06\textwidth}{330} & \renewcommand{\multirowsetup}{\centering} \multirow{11}{0.06\textwidth}{194} & Stacked Active Shape Model (STASM)  \cite{milborrow2008locating} & 11.10 & - \\
& & & & Component-based ASM (ComASM) \cite{le2012interactive} & 9.10 & -\\
& & & & Explicit Shape Regression (ESR) \cite{cao2012face} & 5.70 & 70 (C++)\\
& & & & Robust Cascaded Pose Regression (RCPR) \cite{burgos2013robust} & 6.50 & 6 (Matlab)\\
& & & & Supervised Descent Method (SDM) \cite{xiong2013supervised} & 5.85 & 21 (C++)\\
& & & & Ensemble of Regression Trees (ERT) \cite{kazemi2014one} & 4.9 & 1000 \\
& & & & Local Binary Feature (LBF) \cite{ren2014face} & 5.41 & 200 (C++)\\
& & & & Fast Local Binary Feature (LBF fast) \cite{ren2014face} & 5.80 & 1500 (C++)\\
& & & & Coarse-to-Fine Shape Searching (CFSS) \cite{zhu2015face} & 4.74 & -\\
& & & & CFSS Practical \cite{zhu2015face} & 4.84 & -\\
& & & & cascade Gaussian Process Regression Trees (cGPRT) \cite{lee2015face} & 4.63 & -\\
& & & & & & \\
& & & \renewcommand{\multirowsetup}{\centering} \multirow{8}{0.06\textwidth}{68\tnote{d}} & Tree Structured Part Model (TSPM)  \cite{zhu2012face} & 8.16 & 0.04 (Matlab) \\
& & & & Discriminative Response Map Fitting (DRMF) \cite{asthana2013robust} & 6.70 & 1 (Matlab)\\
& & & & Robust Cascaded Pose Regression (RCPR) \cite{burgos2013robust} & 5.93 & 12 (Matlab)\\
& & & & Supervised Descent Method (SDM) \cite{xiong2013supervised} & 5.67 & 70 (C++)\\
& & & & Gauss-Newton Deformable Part Model (GN-DPM) \cite{tzimiropoulos2014gauss} & 5.69 & 70 \\
& & & & Coarse-to-fine Auto-encoder Networks (CFAN) \cite{zhang2014coarse} & 5.53 & 20 \\
& & & & Coarse-to-Fine Shape Searching (CFSS) \cite{zhu2015face} & 4.63 & -\\
& & & & CFSS Practical \cite{zhu2015face} & 4.72 & -\\
& & & & Deep Cascaded Regression \cite{lai2015deep} & 4.25 & -\\
\Xhline{0.5pt}
\multirow{14}{0.06\textwidth}{300-W \cite{sagonas2013semi}} &  \multirow{14}{0.15\textwidth}{Large variations in illuminations, expressions, poses and occlusion}  & \renewcommand{\multirowsetup}{\centering} \multirow{14}{0.06\textwidth}{689} & \renewcommand{\multirowsetup}{\centering} \multirow{14}{0.06\textwidth}{68} & Tree Structured Part Model (TSPM)  \cite{zhu2012face} & 12.20 & 0.04 (Matlab) \\
& & & & Discriminative Response Map Fitting (DRMF) \cite{asthana2013robust} & 9.10 & 1 (Matlab)\\
& & & & Explicit Shape Regression (ESR) \cite{cao2012face} & 5.28 & 120 (C++)\\
& & & & Robust Cascaded Pose Regression (RCPR) \cite{burgos2013robust} & 8.35 & -\\
& & & & Supervised Descent Method (SDM) \cite{xiong2013supervised} & 7.50 & 70 (C++)\\
& & & & Ensemble of Regression Trees (ERT) \cite{kazemi2014one} & 6.4 & 1000 \\
& & & & Local Binary Feature (LBF) \cite{ren2014face} & 6.32 & 320 (C++)\\
& & & & Fast Local Binary Feature (LBF fast) \cite{ren2014face} & 7.37 & 3100 (C++)\\
& & & & Coarse-to-Fine Shape Searching (CFSS) \cite{zhu2015face} & 5.76 & 25\\
& & & & CFSS Practical \cite{zhu2015face} & 5.99 & 25\\
& & & & cascade Gaussian Process Regression Trees (cGPRT) \cite{lee2015face} & 5.71 & 93\\
& & & & fast cGPRT \cite{lee2015face} & 6.32 & 871\\
& & & & Tasks-Constrained Deep Convolutional Network (TCDCN) \cite{zhang2016learning} & 5.54 & 59\\
& & & & Deep Cascaded Regression (DCR) \cite{lai2015deep} & 5.02 & -\\
& & & & Megvii-Face++ \cite{huang2015coarse} & 4.54 & -\\
\Xhline{0.5pt}
\multirow{12}{0.06\textwidth}{IBUG \cite{sagonas2013semi}} & \multirow{12}{0.15\textwidth}{Extremely large variations in illuminations, expressions, poses and occlusion}  & \renewcommand{\multirowsetup}{\centering} \multirow{12}{0.06\textwidth}{135} & \renewcommand{\multirowsetup}{\centering} \multirow{12}{0.06\textwidth}{68} & Tree Structured Part Model (TSPM)  \cite{zhu2012face} & 18.33 & 0.04 (Matlab) \\
& & & & Discriminative Response Map Fitting (DRMF) \cite{asthana2013robust} & 19.79 & 1 (Matlab)\\
& & & & Explicit Shape Regression (ESR) \cite{cao2012face} & 17.00 & 120 (C++)\\
& & & & Robust Cascaded Pose Regression (RCPR) \cite{burgos2013robust} & 17.26 & -\\
& & & & Supervised Descent Method (SDM) \cite{xiong2013supervised} & 15.40 & 70 (C++)\\
& & & & Local Binary Feature (LBF) \cite{ren2014face} & 11.98 & 320 (C++)\\
& & & & Fast Local Binary Feature (LBF fast) \cite{ren2014face} & 15.50 & 3100 (C++)\\
& & & & Robust Discriminative Hough Voting (RDHV) \cite{jin2016face} & 11.32 & $<$ 1 (Matlab) \\
& & & & Coarse-to-Fine Shape Searching (CFSS) \cite{zhu2015face} & 9.98 & 25\\
& & & & CFSS Practical \cite{zhu2015face} & 10.92 & 25\\
& & & & Tasks-Constrained Deep Convolutional Network (TCDCN) \cite{zhang2016learning} & 8.60 & 59\\
& & & & Deep Cascaded Regression (DCR) \cite{lai2015deep} & 8.42 & -\\
& & & & Megvii-Face++ \cite{huang2015coarse} & 7.46 & -\\
\Xhline{1pt}
\end{tabular}
\begin{tablenotes}
    \tiny
    \item[a] For LFW, the reported performance of \cite{dantone2012real,cao2012face,burgos2013robust} follows the the evaluation procedure proposed in \cite{dantone2012real}, consisting of a ten-fold cross validation using each time 1,500 training images and the rest for testing. In \cite{belhumeur2013localizing}, the model is trained on Columbia's PubFig \cite{kumar2009attribute}, and tested on all 13,233 images of LFW.
    \item[b] Although used by \cite{belhumeur2013localizing}, the 55 point annotation of LFW is not shared.
    \item[c] LFPW is shared by web URLs, but some URLs are no longer valid. So both the training and test images downloaded by other authors are less than the original version (1,100 training images and 300 test images).
    \item[d] LFPW and HELEN are originally annotated with 29 and 194 points respectively, while later Sagonas \emph{et al.} \cite{sagonas2013semi} re-annotate them with 68 points. Some authors reported their performance on the 68 points version of these databases.
\end{tablenotes}
\end{threeparttable}
\end{table*}

\subsubsection{Accuracy} As shown in Table \ref{tab_evaluation}, the localization error on all these databases has been reduced to less than 10\% of the inter-ocular distance by current state-of-the-art. Except for the extremely challenging IBUG database, the best performance on other databases is about 5\% of the inter-ocular distance. To have an intuitive feeling of the extent of localization error, we exemplify the error range of 10\% and 5\% of the inter-ocular distance respectively in Fig. \ref{fig:error_tolerance} (a) and (b). This implies that most of the localized facial points by the state-of-the-art may lie in the error range depicted by the white circles in Fig. \ref{fig:error_tolerance} (a), while on LFPW annotated with 29 points, the mean error range goes to the white circles in Fig. \ref{fig:error_tolerance} (b). Besides the statistics listed in Table \ref{tab_evaluation}, some authors also compared their methods with human beings and reported close to human performance on LFPW \cite{belhumeur2011localizing,burgos2013robust} and LFW \cite{dantone2012real}.

From Table \ref{tab_evaluation}, we can observe that although generative methods (e.g., the GN-DPM \cite{tzimiropoulos2014gauss}) can produce good performance for face alignment \emph{in-the-wild}, discriminative methods, especially those based on cascaded regression \cite{cao2012face,burgos2013robust,xiong2013supervised,ren2014face,kazemi2014one,zhu2015face,lai2015deep,huang2015coarse}, have been playing a dominate role for this task, partially due to recent development of large unconstrained databases. Furthermore, the deep learning-based approach \cite{sun2013deep,zhang2014facial,huang2015coarse,zhang2016learning} have recently emerged as a popular and state-of-the-art method due to their strong feature learning capability, achieving very accurate (even the best) performance on the challenging 300-W and IBUG databases \cite{sagonas2013semi}.

Fig. \ref{fig:evaluation_alignment_examples} shows some extremely challenging cases on IBUG aligned by eight state-of-the-art methods, from which we can observe that large head poses, extreme lighting, and partial occlusions may pose major challenges for many advanced face alignment algorithms, but good results can still be achieved by some state-of-the-art, for example, by the Tasks-Constrained Deep Convolutional Network (TCDCN) method \cite{zhang2016learning}. Furthermore, we find the Fig. \ref{fig:evaluation_alignment_examples} that: (1) Compared to other facial points, the points around the outline of the face are much more difficult to localize, due to the lack of distinctive local texture. (2) As the points around the mouth are heavily dependent on facial expressions, they are more difficult to localize than those points insensitive to facial expressions, such as the points along the eyebrows, outer corners of the eyes, and the nose tips.

Finally, we have to highlight that the accuracy statistics listed in Table \ref{tab_evaluation} may not fully characterize the behavior of these algorithms, since several factors can complicate the assessment. First, even for the same algorithm, different experimental details and programming skills may results in different performance. Secondly, while the number and variety of training examples have a direct effect on the final performance, the training data of some released software is not clear. Thirdly, as pointed by \cite{yang2015empirical}, the performance of many algorithms is sensitive to the face detection variation, but different systems may employ different face detectors. For example, SDM \cite{xiong2013supervised} employs the Viola Jones detector \cite{viola2004robust}, while GN-DPM \cite{tzimiropoulos2014gauss} uses the in-house face detector of the IBUG group.

\subsubsection{Efficiency} Besides accuracy, efficiency is another key performance indicator of face alignment algorithms. In the last column of Table \ref{tab_evaluation}, we report the efficiency of some algorithms, and highlight the implementation types of them (Matlab or C++). In general, the running time listed here is consistent with the algorithm's complexity. For example, algorithms that involves an exhaustive search of local detectors typically have a high time cost \cite{belhumeur2011localizing,zhu2012face,zhou2013exemplar}, while the cascaded regression methods are extremely fast since both the shape-index feature and the stage regression are very efficient to compute \cite{cao2012face,xiong2013supervised,burgos2013robust}. It is worth noting that impressive speed (more than 1,000 FPS for 194 points on HELEN) has been achieved by the Local Binary Feature (LBF) \cite{ren2014face} and Ensemble of Regression Trees (ERT) \cite{kazemi2014one}, using learning-based features.

\section{Conclusion and prospect}\label{sec_conclusion}
Face alignment is an important and essential intermediary step for many face analysis applications. Such a task is extremely challenging in unconstrained environments due to the complexity of facial appearance variations. However, extensive studies on this problem have resulted in a great amount of achievements, especially during the last few years.

In this paper, we have focused on the overall difficulties and challenges in unconstrained environments, and provide a comprehensive and critical survey of the current state of the art in dealing with these challenges. Furthermore, we hope that the practical aspects of face alignment we organized can provide further impetus for high-performance, real-time, real-life face alignment systems. Finally, it is worth mentioning that some closely related problems are deliberately ignored in this paper, such as facial feature tracking in videos \cite{kapoor2002real,ahlberg2001using} and 3D face alignment \cite{vezzetti20123d}, which are also very important in practice.

Despite of many efforts devoted to face alignment during the last two decades, we have to admit that this problem is far from being solved, and several general promising research directions could be suggested.
\begin{itemize}
\item \emph{Challenging databases collection:} Besides new methodologies, another notable development in the field of face alignment has been the collection and annotation of large facial datasets captured \emph{in-the-wild} (cf., Table \ref{tab_evaluation_database}). But even so, we argue that the collection of challenging databases is still important and has the potential to boost the performance of existing methods. This argument can be partially supported by the fact that: the performance of most algorithms on IBUG is inferior to that on other databases such as LFPW and HELEN, as the training set of these algorithms is typically less challenging compared to IBUG.
\item \emph{Feature learning:} One of the holy grails of machine learning is to automate more and more of the feature engineering process \cite{domingos2012few}, i.e., to learn task-specific features in a data-driven manner. In the field of face alignment, many approaches that employ feature learning techniques, including both shallow feature learning \cite{cao2012face,burgos2013robust,ren2014face} and deep learning \cite{sun2013deep,huang2015coarse} methods, have achieved state-of-the-art performances. We believe that, with the assistance of abundant manually labeled images, automatic feature learning techniques can be a powerful weapon for triumphing over various challenges of face alignment in the wild, and deserve the efforts and smarts of researchers.
\item \emph{Multi-task learning:} Multi-task learning aims to improve the generalization performance of multiple related tasks by learning them jointly, which has proven effective in many computer vision problems \cite{yuan2012visual,zhang2013robust}. For face alignment \emph{in-the-wild}, on the one hand, many factors such as pose, expression and occlusion may pose great challenges; while on the other hand, these factors can be considered jointly with face alignment to expect an improvement of robustness. This has been confirmed by the work of \cite{zhang2014facial}, which proposes to exploit the power of multi-task learning under the deep convolutional network architecture, leading to a better performance compared to single task-based deep model. Although some attempts have been proposed, we believe that multi-task learning remains a meaningful and promising direction for face alignment in future.
\end{itemize}

We believe that face alignment \emph{in-the-wild} is a very exciting line of research due to its inherent complexity and wide practical applications, and will draw increasing attention from computer vision, pattern recognition and machine learning.

\section*{Acknowledgment}
This work is partially supported by National Science Foundation of China (61373060), Qing Lan Project, and the Funding of Jiangsu Innovation Program for Graduate Education (KYLX\_0289).

\end{spacing}

\begin{spacing}{0.75}
\begin{multicols}{2}
{\tiny
\bibliographystyle{elsarticle-num}
\bibliography{face_alignment_survey}
}
\end{multicols}
\end{spacing}

\end{document}